\documentclass[conference]{IEEEtran}
\IEEEoverridecommandlockouts
\usepackage{cite}
\usepackage{amsmath,amssymb,amsfonts}
\usepackage{graphicx}
\usepackage{textcomp}
\usepackage{xcolor}
\def\BibTeX{{\rm B\kern-.05em{\sc i\kern-.025em b}\kern-.08em
    T\kern-.1667em\lower.7ex\hbox{E}\kern-.125emX}}

\usepackage[utf8]{inputenc} %
\usepackage[T1]{fontenc}    %
\usepackage{hyperref}       %
\usepackage{url}            %
\usepackage{booktabs}       %
\usepackage{amsfonts}       %
\usepackage{nicefrac}       %
\usepackage{microtype}      %
\usepackage{xcolor}         %

\usepackage{mathtools}
\usepackage{amsmath}
\usepackage{amssymb}
\usepackage{amsthm}
\usepackage[inline]{enumitem}

\usepackage{dsfont}
\usepackage{wrapfig}
\usepackage{multirow}
\usepackage[binary-units]{siunitx}
\usepackage{fbox}

\usepackage[algo2e,ruled,lined,boxed,commentsnumbered, noend, linesnumbered, procnumbered]{algorithm2e}
\usepackage{xurl}

\DeclareMathOperator{\argmin}{argmin}%

\usepackage{graphicx}
\usepackage{subcaption}

\usepackage{pifont}%

\usepackage{ifthen}
\newboolean{showcomments}
\setboolean{showcomments}{true}
\ifthenelse{\boolean{showcomments}}
{ \newcommand{\mynote}[3]{
		\fbox{\bfseries\sffamily\scriptsize#1}
		{\small$\blacktriangleright$\textsf{\emph{\color{#3}{#2}}}$\blacktriangleleft$}}
	\newcommand{\zzz}[1]{{\setlength{\fboxsep}{2pt}\fcolorbox{black}{yellow}{\textsf{\emph{#1}}}}\xspace}}
{ \newcommand{\mynote}[3]{}
	\newcommand{\zzz}[1]{}}

\newif\ifrevision %
\revisiontrue      %

\newcommand{\sys}{\textsc{Facade}\xspace}
\newcommand{\FL}{\ac{FL}\xspace}
\newcommand{\DL}{\ac{DL}\xspace}

\newcommand{\EL}{\ac{EL}\xspace}

\newcommand{\cifar}{CIFAR-10\xspace}
\newcommand{\imagenette}{Imagenette\xspace}
\newcommand{\flickr}{Flickr-Mammals\xspace}

\newcommand{\dpsgd}{{\xspace}\ac{D-PSGD}\xspace}

\newcommand{\deprl}{\textsc{DePRL}\xspace}
\newcommand{\dac}{\textsc{DAC}\xspace}

\usepackage{cleveref}
\usepackage{amsmath,amssymb,amsfonts,amsthm}
\usepackage{physics}
\usepackage{enumitem}
\usepackage{thmtools} 
\usepackage{thm-restate}
\usepackage{bbm}
\theoremstyle{definition}

\theoremstyle{remark}
\newtheorem{remark}{Remark}
\newtheorem{theorem}{Theorem}

\newtheorem{corollary}[theorem]{Corollary}

\newtheorem{assumption}{Assumption}

\allowdisplaybreaks

\newcommand{\cE}{\mathcal{E}}
\newcommand{\cG}{\mathcal{G}}

\crefname{assumption}{assumption}{assumptions}
\Crefname{theorem}{Th.}{Th.}
\Crefname{definition}{Def.}{Def.}
\Crefname{lemma}{Lem.}{Lem.}
\Crefname{equation}{Eq.}{Eq.}
\Crefname{section}{Sec.}{Sec.}
\Crefname{figure}{Fig.}{Fig.}
\Crefname{algorithm}{Alg.}{Alg.}

\providecommand{\iprod}[2]{\ensuremath{\left\langle #1,\,#2  \right\rangle}}

\graphicspath{ {figures/} }
\usepackage{tikz}
\usepackage[eulergreek]{sansmath}
\usepackage{pgfplots}
\usepackage{pgfplotstable}
\usepackage{textcomp}
\usepackage{comment}
\crefname{assumption}{assumption}{assumptions}
\usepackage[abbreviations]{foreign}

\pgfplotsset{compat=newest}
\usepgfplotslibrary{external,units,colorbrewer,groupplots,fillbetween,statistics}
\tikzsetexternalprefix{figures/}
\tikzset{external/mode=list and make}
\usetikzlibrary{patterns,shapes.misc}

\makeatletter
\begingroup\endlinechar=-1\relax
\everyeof{\noexpand}%
\edef\x{\endgroup\def\noexpand\homepath{%
		\@@input|"kpsewhich --var-value=HOME" }}\x
\makeatother

\def\overleafhome{/tmp}
\newcommand{\inputplot}[2]{%
	\ifx\homepath\overleafhome%
	\IfBeginWith{#1}{plots}{\includegraphics{main-figure#2.pdf}}{#1}%
	\else%
	{\sffamily\scriptsize\input{#1}}
	\fi
}

\newcommand{\newgroupwidth}[2]%
{\expandafter\xdef\csname groupwidth#1\endcsname{#2}}

\newcounter{groupwidth}
\newsavebox{\groupwidthbox}
\makeatletter
{\edef\groupnumber{#1}%
	\stepcounter{groupwidth}%
	\@ifundefined{groupwidth\thegroupwidth}{\pgfmathsetlengthmacro{\mywidth}{\linewidth/\groupnumber}}%
	{\expandafter\let\expandafter\mywidth\csname groupwidth\thegroupwidth\endcsname}%
	\begin{lrbox}{\groupwidthbox}%
		\tikzset{/pgfplots/width={\mywidth}}%
		\ignorespaces}%
	{\end{lrbox}%
	\usebox\groupwidthbox
	\pgfmathsetlengthmacro{\mywidth}{\mywidth + (\linewidth - \wd\groupwidthbox)/\groupnumber}
	\immediate\write\@auxout{\string\newgroupwidth{\thegroupwidth}{\mywidth}}}
\makeatother

\usepackage{acronym}
\acrodef{DL}{decentralized learning}
\acrodef{IFCA}{Iterative Federated Clustering Algorithm}
\acrodef{ML}{machine learning}
\acrodef{D-PSGD}{decentralized parallel stochastic gradient descent}
\acrodef{FL}{federated learning}
\acrodef{SGD}{stochastic gradient descent}
\acrodef{IID}{independent and identically distributed}
\acrodef{non-IID}{non independent and identically distributed}
\acrodef{RMSE}{root mean square error}
\acrodef{RMW}{random model walk}
\acrodef{GL}{gossip learning}
\acrodef{EL}{epidemic learning}
\acrodef{DWT}{discrete wavelet transform}
\acrodef{FFT}{fast Fourier transform}
\acrodef{CML}{collaborative machine learning}
\acrodef{EL}{epidemic learning}
\acrodef{IoT}{Internet-of-Things}
\acrodef{DP}{demographic parity}
\acrodef{EO}{equalized odds}
\acrodef{CIFAR}{CIFAR-10}
\acrodef{IMAGENETTE}{Imagenette}
\acrodef{FLICKR}{Flickr-Mammals}
\acrodef{DP}{demographic parity}
\acrodef{EO}{equalized odds}

\title{Fair Decentralized Learning}

\author{
    \IEEEauthorblockN{Sayan Biswas\\EPFL}
    \thanks{This work has been accepted for publication in the IEEE Conference on Secure and Trustworthy Machine Learning (SaTML). The final version will be available on IEEE Xplore.}
    \and
    \IEEEauthorblockN{Anne-Marie Kermarrec\\EPFL}
    \and
    \IEEEauthorblockN{Rishi Sharma\\EPFL}
    \and
    \IEEEauthorblockN{Thibaud Trinca\\EPFL}
    \and
    \IEEEauthorblockN{Martijn de Vos\\EPFL}
}

\begin{document}

\maketitle
\thispagestyle{plain}
\pagestyle{plain}

\begin{abstract}

Decentralized learning (DL) is an emerging approach that enables nodes to collaboratively train a machine learning model without sharing raw data.
In many application domains, such as healthcare, this approach faces challenges due to the high level of heterogeneity in the training data's feature space.
Such feature heterogeneity lowers model utility and negatively impacts fairness, particularly for nodes with under-represented training data.
In this paper, we introduce \textsc{Facade}, a clustering-based DL algorithm specifically designed for fair model training when the training data exhibits several distinct features.
The challenge of \textsc{Facade} is to assign nodes to clusters, one for each feature, based on the similarity in the features of their local data, without requiring individual nodes to know apriori which cluster they belong to.
\textsc{Facade} (1) dynamically assigns nodes to their appropriate clusters over time, and (2) enables nodes to collaboratively train a specialized model for each cluster in a fully decentralized manner.
We theoretically prove the convergence of \textsc{Facade}, implement our algorithm, and compare it against three state-of-the-art baselines.
Our experimental results on three datasets demonstrate the superiority of our approach in terms of model accuracy and fairness compared to all three competitors.
Compared to the best-performing baseline, \textsc{Facade} on the CIFAR-10 dataset also reduces communication costs by 32.3\% to reach a target accuracy when cluster sizes are imbalanced.

\end{abstract}


\section{Introduction}
\label{sec:intro}

\Acf{DL} is a collaborative learning approach that enables nodes to train a \ac{ML} model without sharing their private datasets with other nodes~\cite{lian2017can}.
In each round of \ac{DL}, nodes first locally train their model with their private datasets.
According to a given communication topology, updated local models are then exchanged with \emph{neighbors} and aggregated by each node.
The aggregated model is then used as the starting point for the next round.
This process repeats until model convergence is reached.
Popular \DL algorithms include \Ac{D-PSGD}~\cite{lian2017can}, \Ac{GL}~\cite{ormandi2013gossip}, and \Ac{EL}~\cite{devos2024epidemic}.

Nodes participating in \ac{DL} are likely to own data with differing \emph{feature distributions}~\cite{caldas2018leaf}. 
For example, in a healthcare scenario, feature heterogeneity may arise due to differences in data acquisition, type of medical devices used, or patient demographics~\cite{rieke2020future,nguyen2022federated}.
Consider a network of hospitals that aims to train an advanced tumor detection model with \ac{DL}.
In this scenario, two different brands of scanners are available on the market, and each hospital uses one or the other.
Due to slight differences in image acquisition processes, the feature distributions of the scanned images will vary between hospitals, most likely resulting in feature skew.
For instance, features such as the image quality or contrast between the tumor and surrounding tissue may differ between the two types of scanners.
If one brand of scanners is more widespread, standard \ac{DL} approaches unfairly push the models towards the majority distribution.
As a result, hospitals in the minority distribution exhibit low model utility, leading to diminished \emph{fairness}.

\begin{figure}[b]
    \centering
    \includegraphics[width=.9\linewidth]{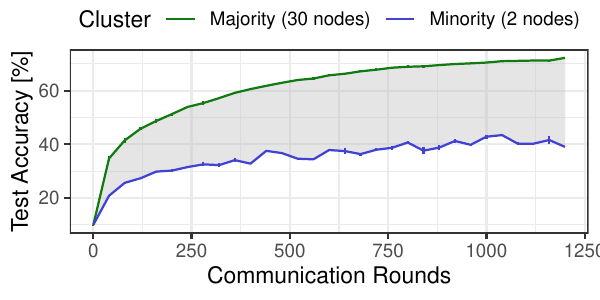}
    \caption{The test accuracy of a model trained with \EL on \cifar. Standard \ac{DL} algorithms such as \dpsgd and \EL results in significantly lower accuracy for the two nodes in the minority cluster compared to that of the nodes in the majority cluster. The error bars indicate the standard deviation of test accuracy.}
    \label{fig:acc_dpsgd_intro}
\end{figure}

We highlight the extent of this fairness issue through an experiment in which we train an image classifier model (LeNet~\cite{lecun1998gradient}) using \Ac{D-PSGD} in a 32-node network on the \cifar dataset~\cite{krizhevsky2009learning}.
For our experiment, we establish two clusters.
The first cluster, the majority cluster, consists of 30 nodes, while the second cluster, the minority cluster, has two nodes.
We ensure a uniform label distribution, with each node having an equal number of samples of each label.
We create a feature distribution shift by turning the \cifar images for the nodes in the minority cluster upside down.
\Cref{fig:acc_dpsgd_intro} shows the averaged test accuracy during training for the nodes in each cluster.
We observe a significant accuracy gap of more than 30\% between the two clusters, demonstrating that the model produced by \dpsgd is much less suitable for nodes in the minority cluster.
This gap is often hidden in results when the global average test accuracy is reported, as this metric is biased towards the majority cluster due to their larger representation.

As \ac{ML}-based decision systems become more prevalent and are increasingly adopted, addressing fairness in \ac{DL} and \ac{ML} as a whole is critical~\cite{pessach2022review}.
These systems must actively mitigate biases that disproportionately harm minority groups~\cite{berk2021fairness}.
In our work, we emphasize that differences in feature distribution should not degrade the prediction quality for nodes that have equally contributed to the learning process.
This is especially crucial in high-stakes domains like healthcare, where the decisions made by \ac{ML} models can have significant real-world consequences for individuals' well-being.

To realize fairness in \ac{DL}, we present \sys (FAir Clustered And Decentralized lEarning), a novel \ac{DL} algorithm designed specifically for settings where training data exhibits two or more underlying features. 
In the presence of such feature heterogeneity, our approach ensures fair predictions across all nodes after model training.
To our knowledge, we are the first to address both model utility and fairness in \ac{DL} settings.

The main idea behind \sys is to group the nodes into clusters based on their feature skew.
The number of clusters is a hyperparameter of \sys, is apriori defined, and should ideally match the number of distinct features in the data.
However, as we will experimentally show in \Cref{sec:exp_estimate_k}, overestimating this number still yields good model accuracy.
To enhance fairness, nodes specialize their models on the feature skew of their cluster while still collaborating with nodes outside their cluster to achieve good model utility across clusters.
This collaboration dynamic allows nodes in a minority cluster to maintain specialized models that perform well on their data while also benefiting from the data of nodes in a majority cluster.
Since the data in \ac{DL} is private and cannot be shared, the key challenge lies in clustering the nodes as part of the \ac{DL} training process in a fully decentralized manner.

\sys accomplishes this by splitting the models into two parts:
\begin{enumerate*}[label=\emph{(\roman*)}]
\item the common collaborative \emph{core}
and
\item a specialized \emph{head}.
\end{enumerate*}
Each node maintains a single core and one head per cluster.
At the start of each training round, each node chooses the head that, combined with the common core, shows the lowest training loss on the current batch of training samples.
The chosen head is then trained along with the common core.
The trained parts of the model are shared and aggregated, similar to \ac{DL} algorithms like \ac{EL}~\cite{devos2024epidemic}.
The dynamic topology used in \ac{EL} provides better mixing of core and heads as the nodes aggregate models with varying nodes.
The clustering process is emergent, \ie, the nodes do not need to know which cluster they belong to.
As the training progresses, nodes with similar feature distributions converge towards the same model head.
This allows the head to specialize in the cluster's feature distribution, resulting in better model utility and improved fairness of predictions across the network.

We implement \sys and conduct an extensive experimental evaluation with three datasets (\cifar, \imagenette and \flickr) and compare \sys to three state-of-the-art baselines (\EL~\cite{devos2024epidemic}, \deprl~\cite{xiong2024deprl} and \dac~\cite{zec2022dac}).
Our evaluation demonstrates that \sys consistently outperforms the baselines in terms of both test accuracy and fairness.
For example, \sys with imbalanced cluster sizes achieves superior test accuracy for minority clusters, with up to 60.0\% on the \cifar dataset, which outperforms the second-best algorithm, \deprl, which reaches 52.6\%.
In addition,  \sys is communication-efficient and requires 41.3\% less communication volume than the \EL baseline to reach a target accuracy of 63\% on \cifar, when cluster sizes are balanced.
Our results underline the strength of \sys in delivering high model utility and fairness while also reducing communication overhead.

\textbf{Contributions.}
Our work makes the following contributions:
\begin{enumerate}
    \item We introduce a novel \ac{DL} algorithm, named \sys, designed to address fairness concerns in the presence of feature heterogeneity (\Cref{sec:design}).
    It ensures high fairness through emergent clustering, based on feature skew, and training specialized models with no additional communication overhead compared to standard \ac{DL} algorithms such as \dpsgd.
    \item We prove the convergence of \sys with an arbitrary number of feature distributions into which the population's data is partitioned (\Cref{sec:theory}). Our bounds depend on the number of nodes, the regularity and smoothness properties of local loss functions, the batch sizes used during local training, and the degree of the random graphs sampled at each communication round.
    \item We experimentally evaluate \sys against three state-of-the-art baselines and datasets (\Cref{sec:eval}).
    Our results demonstrate that \sys results in high model utility for all clusters and excels at maintaining fairness, even for highly imbalanced scenarios where one group significantly outnumbers the other.
\end{enumerate}

\section{Background and Preliminaries}
\label{sec:background}
We first introduce \acl{DL} in~\Cref{subsec:dl} and then elaborate on the notion of fairness used in our work in~\Cref{sec:bg_fairness}.

\subsection{\Acf{DL}}
\label{subsec:dl}
We consider a scenario where a group of nodes, denoted as $\mathcal{N}$, collaboratively train a \ac{ML} model.
This is commonly referred to as \acf{CML}~\cite{soykan2022survey,pasquini2023security}.
Each node $i \in \mathcal{N}$ has its private dataset $D_i$, which is used for computing local model updates.
The data from each node remains on the local device throughout the training process.
The training aims to determine the model parameters $\theta$ that generalize well across the combined local datasets by minimizing the average loss over all nodes in the network.

\Acf{DL}~\cite{nedic2016stochastic,lian2017can,assran2019stochastic} is a class of \ac{CML} algorithms where nodes share model updates with neighboring nodes via a communication topology defined by an undirected graph $\cG=(\mathcal{N},\cE)$.
In this graph, $\mathcal{N}$ is the set of all nodes, and $(i,j) \in \cE$ indicates an edge or communication link between nodes $i$ and $j$.
In standard \DL approaches, $ \cG $ is static but may also vary over time~\cite{lu2023privacy}.
\DL eliminates the need for centralized coordination, enabling each node to collaborate independently with its neighbors.
This decentralized approach enhances robustness against single points of failure and improves scalability compared to alternative \ac{CML} methods like \ac{FL}~\cite{mcmahan2017fedavg}. 

Among the various algorithms, \dpsgd~\cite{lian2017can} is widely regarded as a standard for performing \DL.
In \dpsgd, each node $i$ starts with its loss function $f_i$, initializing its local model $\theta_i^{(0)}$ in the same way before executing the following steps: 

\begin{enumerate}
\item \emph{Local training.} For each round $0\leq t \leq T-1$ and each epoch $0\leq h \leq H-1$, initializing $\tilde \theta_i^{(t,0)}$ as $\theta_i^{(t)}$, node $i$ independently samples $\xi_i$ from its local dataset, computes the stochastic gradient $\nabla f_i(\tilde \theta_i^{(t,h)}, \xi_i)$, and updates its local model by setting $\tilde\theta_i^{(t,h+1)} \leftarrow \tilde\theta_i^{(t,h)} - \eta \nabla f_i(\tilde\theta_i^{(t,h)}, \xi_i)$, where $\eta$ represents the learning rate.
\item \emph{Model exchange.} Node $i$ sends $\tilde\theta_i^{(t,H)}$ to its neighbors and receives $\tilde\theta_j^{(t,H)}$ from each neighboring node $j$ in $\cG$.
\item \emph{Model aggregation.} Node $i$ combines the model parameters received from its neighbors along with its own using a weighted aggregation formula: 
$\theta_i^{(t+1)} = \sum_{\{j: (i,j)\in \cE \} \cup \{i\}} w_{ji} \tilde\theta_j^{\left(t,H\right)}$, where $w_{ji}$ is the $(j,i)^{\operatorname{th}}$ element of the mixing matrix $W$. A typical strategy is to aggregate the models with uniform weights from all neighbors, including the local model.
\end{enumerate}

Upon completing $T$ rounds, node $i$ adopts the final model parameters $\theta_i^{(T)}$, concluding the \ac{DL} training process.
The pseudocode for \dpsgd is included in \Cref{app:dpsgd}.

\subsection{Fairness in Collaborative Machine Learning}
\label{sec:bg_fairness}

With the recent increase in focus on building ethical \ac{ML} models, formally ensuring the fairness of such models during or after training has become an important research challenge~\cite{mehrabi2021survey}.
In the context of \ac{CML}, various methods for quantifying and establishing fairness during model training have been explored from a variety of perspectives~\cite{verma2018fairness, hanna2009measuring, makhlouf2021applicability}. 

\emph{\Ac{DP}}~\cite{dwork2012fairness} is one of the most popular notions of \emph{group fairness} used in \ac{ML} to quantify the presence of bias of any subgroup of the participating data in the outcome of the trained model~\cite{hardt2016equality,ezzeldin2023fairfed,galli2023advancing}. It formally ensures that the different demographic groups present in a population (\eg, based on gender or race) receive positive treatment with equal likelihood, \ie, $\mathbb{P}[\hat{Y}=y|S=0]= \mathbb{P}[\hat{Y}=y|S=1]$ where $\hat{Y}$ denotes the variable representing the decision (label) predicted by the model for a certain input, $y$ is the corresponding positive decision, and $S=0$ and $S=1$ denote the clusters where the sensitive attributes of the input come from (\eg, representing the \emph{majority} and the \emph{minority} in the population). 

To quantify the overall fairness of a \ac{ML} algorithm trained for classification tasks w.r.t. \ac{DP} across all classes, we consider the sum of the absolute difference in \ac{DP} across all possible labels~\cite{denis2024fairness,rouzot2022learning}.
In other words, setting $\mathcal{Y}$ as the space of all possible labels that can be predicted, the overall \ac{DP} across all classes ensured by the algorithm is given as: 
\begin{align}
    &\sum_{y\in \mathcal{Y}}\left\lvert\mathbb{P}[\hat{Y}=y|S=0]-\mathbb{P}[\hat{Y}=y|S=1]\right\rvert.\label{eq:DP_overall}
\end{align}

\emph{\Ac{EO}}~\cite{hardt2016equality} is another widely used metric for group fairness in decision-making algorithms and classification-based \ac{ML}~\cite{galli2023advancing,rouzot2022learning}.
While the high-level goal of the \ac{EO} fairness metric is also to ensure that a model performs equally well for different groups, it is stricter than \ac{DP} because it requires that the model’s predictions are not only independent of sensitive group membership but also that groups have the same true and false positive rates.
This distinction is crucial because a model could achieve \ac{DP}, \ie, its predictions could be independent of sensitive group membership but still generate more false positive predictions for one group versus others.
Formally, an algorithm satisfies \ac{EO} if $\mathbb{P}[\hat{Y}=y|Y=y, S=1] - \mathbb{P}[\hat{Y}=y|Y=y, S=0]$, where, in addition to $\hat{Y}$ and $S$ as described above, $Y$ is the target variable denoting the ground truth label of an associated input.
Similar to measuring the overall \ac{DP} of a classification-based learning algorithm as given by \Cref{eq:DP_overall}, the overall \ac{EO} achieved by the algorithm across all classes is measured as the sum of the absolute differences in the \ac{EO} across every possible label, \ie,

\begin{align}
    \sum_{y\in\mathcal{Y}}\left\lvert\mathbb{P}[\hat{Y}=y|Y, S=1] - \mathbb{P}[\hat{Y}=y|Y, S=0]\right\rvert.\label{eq:EO_overall}
\end{align}

We quantify the fairness guarantees of \sys and baseline approaches by evaluating both the \ac{DP} and \ac{EO} metrics across various learning tasks (see~\Cref{sec:exp_fairness}). Referring to the healthcare example presented in \Cref{sec:intro}, our goal is to ensure that under \sys, patients receive diagnoses of equal quality, \eg, the model predicting whether a tumor is malignant or benign based on medical images, regardless of the brand of scanner used to obtain those images.
In other words, \sys aims to ensure that, under formal standards of \ac{DP} and \ac{EO}, the diagnosis predicted by the model, \ie, $\hat{Y}$ in the above definitions, is independent of any potential bias in the feature distribution of the images.
Such biases may arise due to differences in the technical specifications of the scanners, causing the emergence of majority and minority groups (\ie $S$ in the above definitions).

\begin{figure*}[t!]
    \vspace{-0.4cm}
    \centering
    \includegraphics[width=\textwidth]{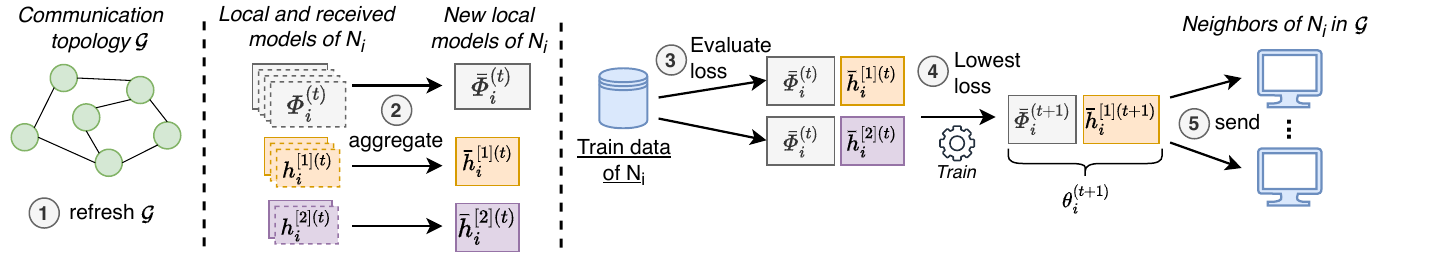}
    \caption{The different operations during a training round in \sys, from the perspective of node $ N_i $.} %
    \label{fig:mechanism}
    \vspace{-0.4cm}
\end{figure*}

\section{Design of \sys}
\label{sec:design}

Our work addresses fairness issues in \ac{DL} due to potential bias in data features among nodes belonging to majority groups.
We design \sys to operate within a permissioned network environment where membership is regulated.
This assumption aligns with the notion that \ac{DL} is often employed in enterprise contexts, where network access is typically restricted~\cite{beltran2023decentralized}.
Examples of such environments include hospitals collaborating on \ac{DL} tasks~\cite{shiranthika2023decentralized}.
In permissioned networks, all nodes are verified entities with established identities.
Thus, threats in open networks, such as Sybil attacks~\cite{douceur2002sybil}, fall outside the scope of this work.
Additionally, we assume that nodes in \sys execute the algorithm correctly, and we consider issues like privacy or poisoning attacks beyond our scope.
In line with related work in this domain, \sys is a synchronous learning system where communication and computations happen in discrete steps and rounds.
Finally, \sys leverages a dynamic topology approach where the communication graph changes each round~\cite{devos2024epidemic}.
Refreshing this topology can be achieved by having nodes participate in topology construction using a decentralized peer-sampling service~\cite{antonov2023securecyclon,guerraoui2024peerswap}.

In the following, we first explain the high-level operations of \sys in~\Cref{subsec:nutshell}.
We then provide a formal, detailed algorithm description in the remaining subsections.

\subsection{\sys in a nutshell}
\label{subsec:nutshell}
We visualize the workflow of \sys in~\Cref{fig:mechanism}.
The underlying idea behind \sys is to maintain $k$ model heads per node, one for each cluster.
Specifically, each model is split into a core and $ k $ heads.
We note that this head-core split is also used by related work on clustering-based \ac{FL}~\cite{Ghosh_2020_IFCA}.
In practice, the model heads are the final few layers of the neural network.
These model heads can then effectively capture the unique variations of features across clusters.
Thus, each node stores $k$ versions of the same head with different parameters, unlike other \DL methods that only keep one model core and head in memory.
\sys is explicitly designed for settings where the number of distinct feature distributions is significantly smaller than the total number of nodes.

At the start of each round, the communication topology is first randomized (step 1 in~\Cref{fig:mechanism}).
Topology randomization in \sys is important to ensure good performance as otherwise nodes could become stuck with sub-optimal neighbors.
Each node $ n $ then receives model cores and heads from its neighbors and aggregates these received model cores and corresponding model heads (step 2).
Specifically, if a node $ n $ receives a head $ h_i $, $ n $ will aggregate $ h_i $ with its own $ i^{th} $ head.
Furthermore, the core is always aggregated.
Each node $ n $ then evaluates the training loss on its mini-batch of training samples, using the model core and for each of the local model heads (step 3).
$ n $ then selects the model head with the lowest training loss and trains it, together with the model core (step 4).
Finally, $ n $ sends the updated model core and head to its neighbors in the communication topology (step 5).
This process repeats until the model has converged.

A key benefit of \sys is that no explicit clustering is required: nodes are not pre-assigned to specific clusters before the training process.
Instead, nodes in \sys are implicitly clustered through dynamic head evaluation and selection.
This procedure allows the clustering to evolve dynamically, enabling nodes to detect similarities with others as models become more specialized and accurate.

\subsection{Notations}\label{sec:notations}

Let $\mathcal{N}=\{N_1,\ldots N_n\}$ be the set of all nodes in the network %
and let $Z_i\subset \mathcal{Z}$ denote the local dataset of $N_i$ for each $i\in [n]$, where $\mathcal{Z}$ is the space of all datapoints. We consider a setting where the features of the data held by each of the $n$ nodes are partitioned into $k$ distinct distributions, $\mathcal{D}_1,\ldots, \mathcal{D}_k$, for some $k \leq n$. %
In other words, we assume that for every $i\in [n]$, there exists a unique $j_i\in [k]$ such that %
the datapoints in $Z_i$ are sampled from $\mathcal{D}_{j_i}$ and we refer to $j_i$ as the \emph{true cluster ID} of $N_i$. For every $j\in [k]$, let $S^*_{[j]}\subseteq \mathcal{N}$ denote the set of nodes whose true cluster identity is $j$, \ie, $S^*_{[j]}=\{N_i\in\mathcal{N}\colon\, Z_i\sim \mathcal{D}_j \}$, and we refer to $S^*_{[j]}$ as the $j$'th \emph{cluster} of the network.

\subsection{Problem formulation}

For each $i\in[n]$, let $f_i:\mathbb{R}^d\mapsto \mathbb{R}$ be the loss function of $N_i$. In practice, nodes sample a finite subset of their personal data set, referred to as a \emph{batch}, for their local training. Let $\xi^{(t)}_i\subseteq Z_i$ represent the batch sampled in round $t$ by $N_i$, independent of the past and other nodes; we assume that the batch size $B=|\xi^{(t)}_i|$ remains constant for all nodes in every round. 
Thus, for every $i\in [n]$, $f_i(\theta,\xi_i^{(t)})=\frac{1}{B}\sum_{z\in \xi^{(t)}_i} f_i(\theta,z)$ is the empirical loss evaluated by $N_i$ on batch $\xi^{(t)}_i$ for model $\theta$. 

We recall that \sys achieves fairness in collaboratively trained models through clustering in a decentralized framework, accounting for the heterogeneous feature distribution of data across nodes.
This is ensured by the fact that \sys trains models tailored for each cluster specific to the feature distribution of their data. Hence, for every $j\in [k]$, the nodes in $S^*_{[j]}$ seek to minimize the average population loss of the $j$'th cluster given by $F^{[j]}=\frac{1}{|S^*_{[j]}|}\sum_{i':N_{i'}\in S^*_{j}}f_{i'}$. 
Thus, the training objective of \sys is to find an optimal model for each cluster that minimizes the corresponding expected population loss, which can be formally expressed as:
\begin{align}
    &\forall\,j\in[k]:\theta^{*}_{[j]}= \argmin_{\theta\in \mathbb{R}^d}F^{[j]}(\theta)\nonumber\\
    &\text{where }F^{[j]}(\theta)=\frac{1}{|S^*_{[j]}|}\sum_{i':N_{i'}\in S^*_{j}}f_{i'}(\theta,Z_{i'})
    \;\;\; \forall j \in [k].\nonumber
\end{align}

\subsection{Detailed algorithm description}
\label{subsec:detailed_algo}

In each round $t$, the nodes furnish a randomized $r$-regular undirected topology $\mathcal{G}_t=(\mathcal{N},\mathcal{E}_t)$ where $\mathcal{E}_t$ is the set of all edges, denoting communication between a pair of nodes, formed independently of the previous rounds. Let $\mathcal{V}_i^{(t)}\subset \mathcal{N}$ be the set of all neighbors of $N_i$ for every $i\in [n]$.  

For every $i\in [n]$, $j\in[k]$, and in any round $t$, let $\theta^{(t)}_i$ be $N_i$'s local model divided into a \emph{head} $h^{(t)}_i$ and a \emph{core} $\phi^{(t)}_i$ and we write $\theta^{(t)}_i=h^{(t)}_i\circ \phi^{(t)}_i$. In round $t=0$, \sys starts by initializing $k$ models that are shared with every node, where each model consists of a cluster-specific unique head and a common core.
During each round, every node receives models (divided into head and core) from their neighbors, aggregates all the cores with its own, and performs a cluster-wise aggregation of the heads. Each node then concatenates the aggregated core with the $k$ (cluster-wise) aggregated heads to form $k$ different models and identifies the one that results in the least loss with its own data. It then performs the SGD steps on that model before sharing the locally optimized model and its corresponding cluster ID with its neighbors.

For every $i\in [n]$, $j\in[k]$, and in any round $t$, set $S^{(t)}_{[j]}\subseteq \mathcal{N}$ to be the set of all nodes reporting their cluster ID as $j$ in round $t$. Then, let $\bar{\theta}^{[j](t)}$ be the global aggregated model for the $j$'th cluster given by $\bar{\theta}^{[j](t)}=\frac{1}{|S^{(t)}_{[j]}|}\sum_{i\in S^{(t)}_{[j]}}\theta^{(t)}_i$ and let $\bar{\theta}_i^{[j](t)}=\bar{h}^{[j](t)}_i\circ\bar{\phi}^{(t)}_i$ be the aggregate of the models with the cluster identity $j$ received by $N_i$ where $h^{[j](t)}_i$ and $\phi^{(t)}_i$ are computed as:
\begin{align}
    &\bar{\phi}^{(t)}_i=\frac{1}{|\hat{\mathcal{V}}_i^{(t)}|}\sum_{i'\in \hat{\mathcal{V}}_i^{(t)}}\phi_{i'}^{(t)}\text{ and }\label{eq:cluster_core_agg}\\
    &\bar{h}^{[j](t)}_i=\frac{1}{|S^{(t)}_{j}\cap \hat{\mathcal{V}}_{i}^{(t)}|}\sum_{i':N_{i'}\in S^{(t)}_{j}\cap \hat{\mathcal{V}}_i^{(t)}}h^{(t)}_{i'},\label{eq:cluster_head_agg}\\
    &\text{where $\hat{\mathcal{V}}_{i}^{(t)}=\mathcal{V}_{i}^{(t)}\cup \{N_i\}$.}\nonumber
\end{align}

\setlength{\fboxsep}{5pt} %

The workflow of the \sys algorithm can formally be described as follows.

\vspace{10pt}
\noindent
\fparbox{
\vspace{2pt}
\begin{center}
 \textbf{The \sys Algorithm} \\[7pt]   
\end{center} 

Round $T \geq t \geq 0$ in \sys consists of the following steps.

\begin{enumerate}
    \item \emph{Randomized topology.}
    The randomized communication topology $\mathcal{G}=(\mathcal{N},\mathcal{E}_t)$ is established. 
    \item \emph{Local steps.} 
    For all $i\in[n]$ and in any round $t$:
    \begin{enumerate}
        \item \emph{Receive models.}
        $N_i$ receives $\theta_{i'}^{(t)}=h_i^{(t)}\circ \phi_i^{(t)}$ and the corresponding cluster IDs from each $i'\in \mathcal{V}_i^{(t)}$.
        \item \emph{Cluster-wise aggregation.} $N_i$ aggregates the cores and performs a cluster-wise aggregation of the heads as given by \Cref{eq:cluster_core_agg,eq:cluster_head_agg}, respectively, to obtain $\bar{\theta}_i^{[j](t)}=\bar{h}_i^{(t)}\circ \bar{\phi}_i^{(t)}$ for every cluster $j\in[k]$. %
        \item \emph{Cluster identification.} $N_i$ obtains the cluster ID that gives the least loss on its local data. 
        $$\hat{j}^{(t)}_i=\argmin_{j\in [k]}\nabla f_i (\bar{\theta}_i^{[j](t)},\xi^{(t)}_i)$$
        \item \emph{Local training.}
         Let $\tilde \theta_i^{[\hat{j}_i^{(t)}](t,0)}=\bar{\theta}_i^{[\hat{j}_i^{(t)}](t)}$. 
    \begin{align}
    &\text{For } h\leq H-1\colon\nonumber\\
        &\tilde \theta_i^{[\hat{j}_i^{(t)}](t,h+1)}=\tilde \theta_i^{[\hat{j}_i^{(t)}](t,h)}-\eta \nabla f_i (\tilde\theta_i^{[\hat{j}_i^{(t)}](t,h)},\xi^{(t)}_i),\nonumber
    \end{align}
    where $\nabla f_i(\tilde\theta_i^{[\hat{j}_i^{(t)}](t,h)},\xi^{(t)}_i)$ is the stochastic gradient of $f_i$ computed on batch $\xi^{(t)}_i$, $\eta>0$ is the learning rate, and $H$ is the number of local SGD steps performed.
    \end{enumerate}
    \item \textit{Communication.} 
    For each $i\in [n]$, set $\theta_i^{(t+1)}=\tilde\theta_i^{[\hat{j}_i^{(t)}](t,H)}$ and share $(\theta_i^{(t+1)}, \hat{j}^{(t)}_i)$ with nodes in $\mathcal{V}_i^{(t)}$.
\end{enumerate}
}
\vspace{5pt}

\subsection{Discussion}
We now discuss various properties of the \sys algorithm.

\textbf{Random Topologies.}
We randomize the communication topology for the following two reasons.
First, it has been demonstrated that altering random communication topologies leads to faster model convergence than traditional \DL approaches that maintain a static, fixed topology throughout training~\cite{devos2024epidemic}. 
Second, in \sys, dynamic topologies also prevent nodes in a cluster from becoming isolated due to initial neighbors from other clusters.
By sampling random neighbors each round, an isolated node will likely eventually exchange models with nodes with similar data features.

\textbf{Overhead.}
Compared to \dpsgd, \sys incurs some storage and compute overhead.
The storage overhead stems from each node having to store $ k $ model heads.
However, since model heads generally consist of a very small number of parameters compared to the full model, the storage overhead of storing $ k $ model heads is negligible.
Furthermore, there is additional compute overhead as each node is required to compute $ k $ training losses each training round, one for each model head.
However, this overhead is also manageable since one can store the output tokens of the model core and input these to each model head.
Alternatively, the $ k $ forward passes can also be computed in parallel.

\textbf{Choice of $ k $.}
As with many other clustering algorithms~\cite{macqueen1967some,reynolds2009gaussian}, the number of clusters ($k$) is a hyperparameter, which should be estimated by the system designer beforehand.
This value heavily depends on the application domain and characteristics of individual training datasets.
We experimentally show in~\Cref{sec:exp_estimate_k} that our algorithm performs well even if the chosen $ k $ is not exactly equal to the true number of feature distributions in the network.

\section{Theoretical analysis of convergence}
\label{sec:theory}
We now theoretically analyze the convergence of the cluster-wise aggregated models under \sys. We begin by introducing some additional notation and outlining the assumptions we make, followed by deriving the intermediate results (\Cref{th:local_convergence,th:global_convergence}) that lead to the final result (\Cref{th:final_convergence}).

For every cluster $j\in[k]$, we assume the existence of an optimal model $\theta^{*}_{[j]}$ that minimizes the average loss of the nodes in $S^{*}_{[j]}$. In order to theoretically analyze the convergence of \sys, in addition to the notations introduced in \Cref{sec:notations}, we introduce the following additional terms.
For every $i\in[n]$, $j\in[k]$, and in any round $t$, let $F^{[j](t)}_i=\frac{1}{|S^{(t)}_{j}\cap \hat{\mathcal{V}}_{i}^{(t)}|}\sum_{i'\in S^{(t)}_{j}\cap \hat{\mathcal{V}}_{i}^{(t)}} f_{i'}$ be the average loss of the neighbors of $N_i$ who report their cluster ID as $j$ in round $t$ and we refer to this quantity as the \emph{local population loss} in the for the $j$'th cluster in the neighborhood of $N_i$ in round $t$.
Finally, for every $j\in [k]$, let $p_{[j]}=\frac{|S^{*}_{j}|}{n}$ denote the fraction of nodes belonging to cluster $j$ with $p=\min_{j\in[k]}\{p_{[j]}\}$ and let $\Delta$ denote the \emph{minimum separation} between the optimal models of every cluster, \ie, $\Delta=\min_{j\neq j'}\norm{\theta^*_{[j]}-\theta^*_{[j']}}$.

We now proceed to study how the cluster identities reported by each node in every round of \sys correspond to their true cluster identities over time. %
 In order to develop the theoretical analysis, we adhere to a standard set of assumptions which are widespread in related works~\cite{Ghosh_2020_IFCA,cyffers2022muffliato,devos2024epidemic,biswas2024beyond,biswas2024lowcost}. In the interest of space, the proofs of all the theoretical results presented in this section are postponed to \Cref{app:proofs}.

\begin{assumption}\label{assump:strong_convex_L_smooth}
    For every $i\in [n]$, $j\in [k]$, and in round $t\geq 0$, the corresponding local population loss of the $j^{th}$ cluster in the neighborhood of $N_i$ in round $t$ is $\lambda$-convex and $L$-smooth. Formally, for every $\theta,\theta'\in \mathbb{R}^d$, there exist constants  $\lambda\geq 0$ and $L\geq 0$ such that:
    \begin{align}
        &F^{[j](t)}_i(\theta')-F^{[j](t)}_i(\theta)\nonumber\\ 
        &\geq \iprod{\nabla F^{[j](t)}_i(\theta)}{\theta'-\theta} + \frac{\lambda\norm{\theta'-\theta}_2^2}{2}\text{ ($\lambda$-convexity) and}\nonumber\\
        &\norm{\nabla F^{[j](t)}_i(\theta)- \nabla F^{[j](t)}_i(\theta')}\leq L \norm{\theta-\theta'} \text{ ($L$-smoothness)}.\nonumber
    \end{align}
\end{assumption}

\begin{assumption}\label{assump:bounded_var}
    For every $i\in[n]$ and in any round $t\geq0$, the variances of $f_i$ and $\nabla f_i$ computed on batch $\xi_i^{(t)}\subset Z_i$, sampled according to $\mathcal{D}_{j_i}$, are bounded by $\sigma^2$ and $\nu^2$, respectively, \ie
    \begin{align}
        &\mathbb{E}_{\xi_i\sim \mathcal{D}_{j_i}}\left[\left(f_i(\theta,\xi_i^{(t)})-F^{[j_i](t)}_i(\theta)\right)^2\right]\leq \sigma^2\text{ and}\nonumber\\
        &\mathbb{E}_{\xi_i^{(t)}\sim \mathcal{D}_{j_i}}\left[\norm{\nabla f_i(\theta,\xi_i^{(t)})-\nabla F^{[j_i](t)}_i(\theta)}^2\right]\leq \nu^2.\nonumber
    \end{align}
\end{assumption}

In addition to Assumptions~\ref{assump:strong_convex_L_smooth} and \ref{assump:bounded_var}, we also assume some initialization conditions of \sys. These are consistent with the standard works on clustering-based personalized model training in \ac{FL}~\cite{Ghosh_2020_IFCA}.

\begin{assumption}\label{assump:init}
    For every cluster $j\in[k]$, we assume \sys to satisfy the following initialization conditions:
    \begin{align}
         &\norm{\bar{\theta}_i^{[j](0)}-\theta^{*}_{[j]}}\leq (\frac{1}{2}-\alpha)\sqrt{\frac{\lambda}{L}}\Delta ,\nonumber\\
         & \text{where $0\leq\alpha \leq\frac{1}{2}$},\,B\geq \frac{k \sigma^2}{\alpha^2\lambda^2\Delta^4},\, p\geq \frac{\log(nB)}{B},
         \text{ and}\nonumber\\
         &\Delta \geq \Bar{\mathcal{O}}\left(\max\{\alpha^{-2/5}B^{-1/5},\,\alpha^{-1/3}n^{-1/6}B^{-1/3}\}\right).\nonumber
     \end{align}
\end{assumption}

 \begin{theorem}\label{th:local_convergence}
     If \Cref{assump:strong_convex_L_smooth,assump:bounded_var,assump:init} hold, choosing learning rate $\eta=1/L$, for a fixed node $N_i$, each cluster $j\in [k]$, and any $\delta\in(0,1)$, in every round $t>0$, we have with probability at least $(1-\delta)$:
     \begin{align}
         \norm{\bar{\theta}_i^{[j](t+1)}-\theta^{*}_{[j]}}\leq (1-\frac{p\lambda}{8L}) \norm{\bar{\theta}_i^{[j](t)}-\theta^{*}_{[j]}}+\epsilon_0\nonumber
     \end{align}
     where $\epsilon_0 \leq \frac{\nu}{\delta L \sqrt{pnB}}+\frac{\sigma^2}{\delta \alpha^2 \lambda^2 \Delta^4 B}+\frac{\sigma \nu k^{3/2}}{\delta^{3/2}\alpha \lambda L \Delta^2 \sqrt{n}B}$.
 \end{theorem}

  \begin{theorem}\label{th:global_convergence}
     If \Cref{assump:strong_convex_L_smooth,assump:bounded_var,assump:init} hold, choosing learning rate $\eta=1/L$, for each cluster $j\in [k]$ and in every round $t>0$, let %
     the locally aggregated model for any cluster at each node be independent of each other. Then for any $\delta\in (0,1)$, we have with probability at least $(1-\delta)^{|S^{(t)}_{[j]}|}$:
     \begin{align}
         &\norm{\bar{\theta}^{[j](t+1)}-\theta^{*}_{[j]}}\leq (1-\frac{p\lambda}{8L}) \norm{\bar{\theta}^{[j](t)}-\theta^{*}_{[j]}}+\epsilon_0\nonumber
     \end{align}
     where $\epsilon_0 \leq \frac{\nu}{\delta L \sqrt{pnB}}+\frac{\sigma^2}{\delta \alpha^2 \lambda^2 \Delta^4 B}+\frac{\sigma \nu k^{3/2}}{\delta^{3/2}\alpha \lambda L \Delta^2 \sqrt{n}B}$.
 \end{theorem}

Finally, in order to derive the convergence analysis for each cluster, in every round $t>0$ and for each cluster $j\in [k]$, let the global aggregate model for cluster $j$ be given by $\theta^{[j](t)}=\frac{1}{|S^{(t)}_{[j]}|}\sum_{i:N_i\in S^{(t)}_{[j]}}\bar{\theta}_i^{[j](t)}$. The following result now helps us understand the cluster-wise convergence of the models under \sys.  

 \begin{corollary}\label{th:final_convergence}
     If \Cref{assump:strong_convex_L_smooth,assump:bounded_var,assump:init} hold, choosing learning rate $\eta=1/L$, for each cluster $j\in [k]$ and in every round $t>0$, let the locally aggregated model for any cluster at each node be pairwise independent of each other. Then for any $\delta\in (0,1)$ and any $\epsilon>0$, setting $\hat T=\frac{8L}{p\lambda}\log(\frac{2\Delta}{\epsilon})$, in any round $t\geq \hat T$ of \sys, we have with probability at least $(1-\delta)^{|S^{(t)}_{[j]}|}$:
     \begin{align}
         &\norm{\theta^{[j](t)}-\theta^{*}_{[j]}}\leq\epsilon\nonumber
     \end{align}
     where $\epsilon \leq \frac{\nu k L \log(nB)}{p^{5/2}\lambda^2 \delta \sqrt{nB}}+\frac{\sigma^2 L^2 k \log(nB)}{p^2 \lambda^4 \delta \Delta^4 B}+\Bar{\mathcal{O}}(\frac{1}{B\sqrt{n}})$.
 \end{corollary}

\begin{remark}
    For any $i \in [n]$ and in each round $t$, as the degree of $N_i$ in the communication topology $\mathcal{G}_t$ increases, $F^{[j_i](t)}_i$ provides a better estimate of $F^{[j_i]}$, which consequently reduces the upper bounds of the variances of $f_i$ and $\nabla f_i$ (\ie, $\sigma^2$ and $\nu^2$, respectively) under \Cref{assump:bounded_var}. This occurs because the local population loss in the neighborhood of $N_i$ for its true cluster ID $j_i$ becomes closer to the overall population loss for cluster $j_i$. Hence, from \Cref{th:final_convergence}, observing that the convergence rates of the cluster-wise aggregated models are directly proportional to $\sigma^2$ and $\nu^2$, we conclude that as the degree of regularity of the communication topology increases, we expect faster convergence, with the extreme case of a fully connected topology converging at the same rate as in \ac{FL}.

\end{remark}

\begin{remark}
\Cref{th:global_convergence} implies that the cluster-wise aggregate of the models received by each node from its neighbors get probabilistically and progressively closer to the optimal model for the corresponding cluster between any two \emph{consecutive} rounds up to a small bounded error of $\epsilon_0$. %
This serves as an intermediate step in deriving the final convergence guarantee of \sys given by \Cref{th:final_convergence}.
\Cref{th:final_convergence} shows that for any error term $\epsilon$, there exists a round $\hat T$ such that, for every round $t \geq \hat T$, the difference between the cluster-wise aggregated model over the entire population and the optimal model for the corresponding cluster probabilistically becomes smaller than $\epsilon$. The bounded term $\epsilon$ can be interpreted as a \emph{precision tolerance} for the convergence of \sys, which may be adjusted according to the application-specific requirements. It is important to note that this precision tolerance $\epsilon$ and the time required to reach the corresponding level of precision $\hat{T}$ are inversely proportional to each other. For instance, in cases where extremely precise convergence is not necessary (\ie $\epsilon$ is large), the number of rounds required to achieve the desired convergence will be relatively small. Conversely, in scenarios where highly accurate convergence is critical (\ie, $\epsilon$ is sufficiently small), the number of rounds required to ensure this precision will be higher. 
\end{remark}%

\section{Experimental Evaluation}
\label{sec:eval}

We conduct an extensive experimental evaluation of \sys and explore its performance against related state-of-the-art \DL algorithms.
We implement \sys and baselines using the \textsc{DecentralizePy} framework~\cite{dhasade2023decentralized} and open-source our code.\footnote{See \url{https://github.com/sacs-epfl/facade}.}

Our experiments answer the following six questions:
\begin{enumerate}
    \item What is the per-cluster test accuracy for \sys and baselines when varying the cluster sizes (\Cref{sec:exp_test_accuracy})?
    \item What is the fair accuracy for \sys and baselines when varying the cluster sizes (\Cref{sec:exp_fair_accuracy})?
    \item What is the fairness for \sys and baselines, in terms of \ac{DP} and \ac{EO}, when varying the cluster sizes (\Cref{sec:exp_fairness})?
    \item How does the communication cost to reach a fixed target accuracy of \sys compare to that of baselines (\Cref{sec:exp_communication_cost})?
    \item What is the impact of wrongly estimating $ k $ (\Cref{sec:exp_estimate_k})?
    \item How does \sys implicitly assign nodes to clusters (\Cref{subsec:exp_settlement})?
\end{enumerate}

\subsection{Experimental setup}

\textbf{Datasets and Partitioning.}
Our experiments focus on supervised image classification as a universal task setting, which is in line with our baselines~\cite{devos2024epidemic,xiong2024deprl,zec2022dac} and related work~\cite{Ghosh_2020_IFCA}.
Specifically, we conduct our experiments on the \cifar~\cite{krizhevsky2009learning}, \flickr~\cite{hsieh2020non} and \imagenette~\cite{imagenette} datasets.
The \cifar dataset is a widely-used image classification task, containing \num{50000} images evenly divided among ten classes (with a resolution of 32x32 pixels).
\imagenette is a subset of ten classes taken from Imagenet~\cite{deng2009imagenet} with \num{9469} images (with a resolution of 224x224 pixels) in total.
Finally, the \flickr dataset contains \num{48158} images of \num{41} different mammal species with varying resolution.
\Cref{table:experiment_datasets} in~\Cref{app:expt_setup} summarizes the used dataset and learning parameters.

To create an environment with clustered data with feature skew, we first uniformly partition each dataset into several smaller subsets, \ie, each client has the same number of training samples from each class.
We use uniform partitioning because our work aims to study fairness in networks with feature heterogeneity, not label skewness.
Thus, the heterogeneity must be reflected in the feature composition of each cluster.
To ensure feature heterogeneity, we randomly apply different rotations to the images of each cluster, ensuring that no two clusters share the same rotation.
Rotation preserves the underlying characteristics of the images and is commonly used by related work~\cite{zec2022dac,Ghosh_2020_IFCA,onoszko2021decentralized,chung2022federated}.
We note that this process maintains the same label distribution across clusters while introducing recognizable differences in feature compositions.
We also experiment with feature heterogeneity by applying color filters to training images, see~\Cref{app:exp_color_shift}.
Additionally, all nodes within the same cluster share a common test set with the same rotation as their training set, ensuring the test data has the same feature shift.

\textbf{Cluster Configurations.}
To assess the fairness of \sys and baselines, we design experiments with multiple clusters with varying proportions. 
We keep the total number of nodes constant for all experiments while varying the cluster sizes.
We demonstrate that, even when the \emph{minority} group is heavily outnumbered by the \emph{majority} group, \sys maintains a high accuracy for all clusters and significantly improves accuracy for the minority group compared to baseline algorithms.

Unless specified otherwise, we experiment with two clusters of varying sizes.
For \cifar, we consider three cluster configurations and \num{32} nodes, with majority-to-minority ratios of 16:16, 24:8, and 30:2.
For instance, for the 24:8 cluster configuration, 24 nodes have upright images, while the remaining \num{8} nodes possess images that are rotated 180°.
For \imagenette, we have \num{24} nodes in total and consider cluster configurations with ratios of 12:12, 16:8, and 20:4.
For \flickr, we have \num{16} nodes and consider two cluster configurations with ratios 8:8 and 14:2.
While most of our experiments focus on a two-cluster setup, \sys can be applied to configurations with more than two clusters, as shown in~\Cref{sec:exp_estimate_k}.
Moreover, our convergence analysis (see~\Cref{sec:theory}) is carried out for an arbitrary number of clusters in the population.

\textbf{Models.}
We use GN-LeNet for the \cifar experiments~\cite{hsieh2020non}.
It has about 120k parameters, consisting of three convolution layers and one feed-forward layer.
When training models with \sys, we designate the last fully connected layer as the head and use the rest of the model as the common core.
We also use a LeNet model for \imagenette but adjust the model to accept images with a 64x64 pixel resolution.
The resulting number of parameters of this model is about 250k.
For \flickr, we use a ResNet8~\cite{he2016deep} model, with roughly 310k parameters, that accept 64x64 pixel images as input.
Since this dataset is more challenging, we modify the head size of ResNet8 and include the last two basic blocks in the head, along with the final fully connected layer.

\textbf{Baselines.}
We compare \sys against three related \ac{DL} algorithms: \EL~\cite{devos2024epidemic}, \deprl~\cite{xiong2024deprl} and \dac~\cite{zec2022dac}.
\EL is a state-of-the-art \ac{DL} algorithm, based on \dpsgd, that leverages randomized communication to improve model convergence.
We have included it as a baseline since \emph{(i)} \sys also relies on communication with random nodes~\cite{devos2024epidemic}, and \emph{(ii)} \dpsgd is a widely used baseline in this domain.

\deprl is a state-of-the-art personalized \ac{DL} algorithm that allows each node to optimize its model head locally while sharing the core model through periodic communication and aggregation steps~\cite{xiong2024deprl}.
In contrast to \sys, \deprl uses a static communication topology and does not share the head with other nodes each communication round.
Furthermore, \deprl focuses on dealing with label heterogeneity, whereas the focus of \sys is on feature heterogeneity instead.

We also compare against \dac, a state-of-the-art approach for personalized \ac{DL} on clustered non-IID data~\cite{zec2022dac}.
This approach adapts the communication weights between nodes based on their data distributions, enhancing learning efficiency and performance in clustered settings.
Like \sys, \dac utilizes a dynamic communication topology and has been tested on cluster configurations similar to our work.
Finally, we remark that none of the selected \ac{DL} baselines consider fairness, which is a unique contribution of our work.

\textbf{Learning Parameters.}
We run each experiment for $T=1200$, $ T = 800 $, and $ T = 1200 $ communication rounds for the \cifar, \imagenette, and \flickr datasets, respectively.
Each local training round features $\tau=10$ local steps with batch size $B=8$.
For \flickr, we increase the number of local steps to $\tau=40$.
We use the SGD optimizer for \sys and all baselines and have independently tuned the learning rate for each learning task and baseline with a grid search.
\Cref{table:experiment_datasets} summarizes these parameters.
For \sys and baselines, we perform an all-reduce step in the final round~\cite{patarasuk2009bandwidth}, where all nodes share their models with everyone else and perform a final aggregation.
Finally, we fix the communication topology degree to \num{4} for \sys and baselines.

To evaluate the algorithms, we measure the test accuracy on the provided test sets every \num{80} rounds.
We also compute the relevant fairness metrics (\ac{DP} and \ac{EO}) of the final models with~\Cref{eq:DP_overall} and~\Cref{eq:EO_overall}.
In the main text, we provide experimental results for the \cifar and \imagenette datasets and complementary experimental results and plots for the \flickr dataset in~\Cref{app:flickr_results}.
We also provide experiments with label skewness in~\Cref{app:label_heterogeneity}.

\begin{figure}[t]
	\centering
	\begin{subfigure}{\columnwidth}
		\centering
		\includegraphics[width=.8\columnwidth]{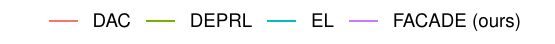}
	\end{subfigure}
	\begin{subfigure}{\columnwidth}
		\centering
		\includegraphics[width=\columnwidth]{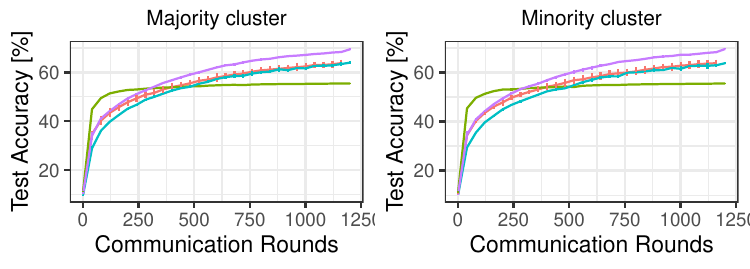}
		\caption{16:16 cluster configuration}
		\label{fig:acc_per_cluster_cifar10_16_16}
	\end{subfigure}
	\begin{subfigure}{\columnwidth}
		\centering
		\includegraphics[width=\columnwidth]{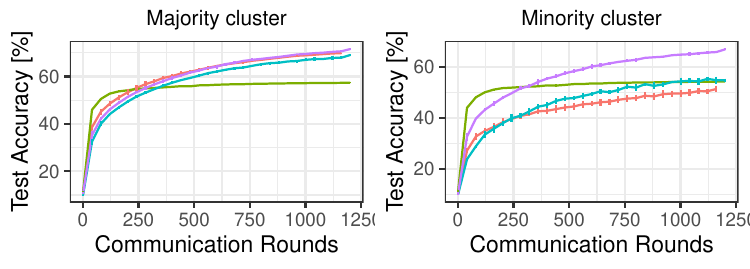}
		\caption{24:8 cluster configuration}
		\label{fig:acc_per_cluster_cifar10_24_8}
	\end{subfigure}
	\begin{subfigure}{\columnwidth}
		\centering
		\includegraphics[width=\columnwidth]{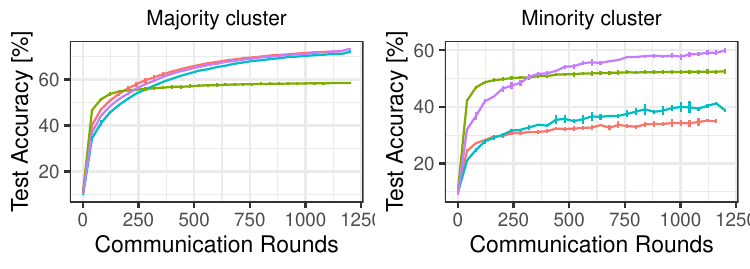}
		\caption{30:2 cluster configuration}
		\label{fig:acc_per_cluster_cifar10_30_2}
	\end{subfigure}%
	\caption{Average test accuracy for the nodes in the majority cluster (left) and those in the minority (right) obtained on \cifar ($\uparrow$ is better), for different cluster configurations.}
	\label{fig:acc_clust_cifar}
\end{figure}

\begin{figure}[t]
	\centering
	\begin{subfigure}{\columnwidth}
		\centering
		\includegraphics[width=.8\columnwidth]{figures/acc_per_cluster_legend.pdf}
	\end{subfigure}
	\begin{subfigure}{\columnwidth}
		\centering
		\includegraphics[width=\columnwidth]{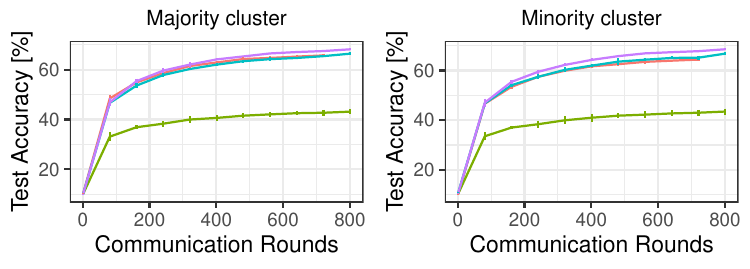}
		\caption{12:12 cluster configuration}
		\label{fig:acc_per_cluster_imagenette_12_12}
	\end{subfigure}
	\begin{subfigure}{\columnwidth}
		\centering
		\includegraphics[width=\columnwidth]{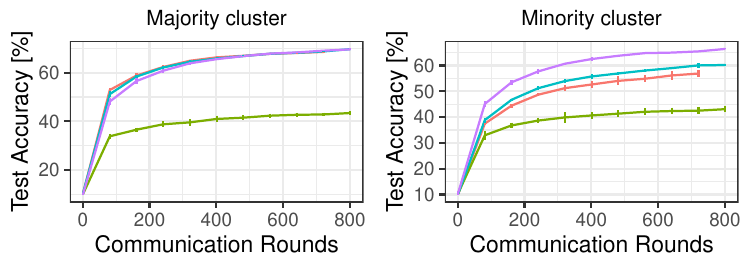}
		\caption{16:8 cluster configuration}
		\label{fig:acc_per_cluster_imagenette_16_8}
	\end{subfigure}
	\begin{subfigure}{\columnwidth}
		\centering
		\includegraphics[width=\columnwidth]{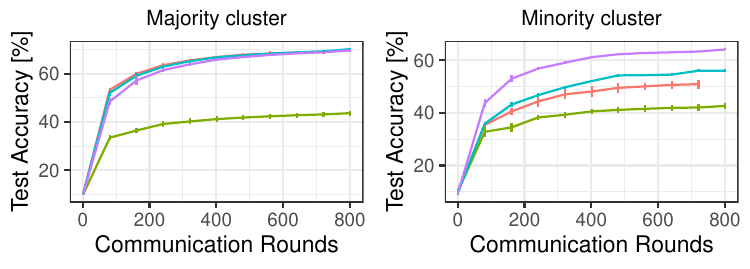}
		\caption{20:4 cluster configuration}
		\label{fig:acc_per_cluster_imagenette_20_4}
	\end{subfigure}%
	\caption{Average test accuracy for the nodes in the majority cluster (left) and those in the minority (right) obtained on \imagenette ($\uparrow$ is better), for different cluster configurations.}
	\label{fig:acc_clust_imagenette}
\end{figure}

\subsection{Per-cluster test accuracy for varying cluster configurations}
\label{sec:exp_test_accuracy}
We analyze the per-cluster test accuracy for the \cifar and \imagenette datasets when varying the cluster size.
Our goal is to analyze the performance of \sys and baselines when the minority group is increasingly outnumbered.

\Cref{fig:acc_clust_cifar} shows the average test accuracy for each cluster for the \cifar dataset as model training progresses for the three considered cluster configurations.
The test accuracy of the majority and minority clusters is shown in the left and right columns, respectively.
When both clusters are of equal size (\Cref{fig:acc_per_cluster_cifar10_16_16}), \sys attains higher model utility compared to baselines: 64.0\% and 69.5\% for \ac{EL} and \sys, respectively, after $ T = 1200 $.
We attribute this gain to the management of multiple heads by \sys, which provides a greater capacity to adapt to variations in features across clusters.
Generally, most of our baselines show reasonable performance with test accuracies around 63\%, since there is enough data to find a good model that suits both clusters.
We also observe that the attained accuracy of \deprl plateaus early in the training process.
For all cluster configurations, we observe that \sys either outperforms or is at par with the baselines.

When considering the test accuracy of minority clusters (right column of~\Cref{fig:acc_clust_cifar}), we note that \sys consistently outperforms baselines and gives the \textit{minority} groups a test accuracy that, for a fixed cluster configuration, is comparable to the one for the nodes in the majority cluster.
When cluster configurations become more skewed (\Cref{fig:acc_per_cluster_cifar10_24_8} and~\Cref{fig:acc_per_cluster_cifar10_30_2}), the attained test accuracy of \ac{EL} drops significantly.
This is because consensus-based methods like \EL optimize for the data distribution of the majority cluster.
Compared to other baselines, \deprl performs well in the 30:2 cluster configuration (\Cref{fig:acc_per_cluster_cifar10_30_2}) and reaches 55.4\% test accuracy after $ T = 1200 $.
We attribute this performance to the specialization of model heads to the heterogeneous features.
Nevertheless, \sys outperforms all baselines in terms of test accuracy of the minority cluster.
In the 24:8 cluster configuration, \sys reaches 66.8\% test accuracy compared to 54.8\% test accuracy for \EL.
For the highly skewed cluster configuration of 30:2, \sys reaches 60.0\% test accuracy compared to 52.5\% test accuracy for \deprl.
\sys thus demonstrates fair treatment to the minority group.
The superior performance of \sys for the majority group is because the model heads are specialized for each cluster.
This allows the model head to remain unaffected by the majority's data distribution and to adapt specifically to the minority group's data distribution.
We also notice that for all cluster configurations baselines achieve higher accuracy on the majority group, which we believe is because this group has more data samples.

\Cref{fig:acc_clust_imagenette} shows similar plots for the \imagenette dataset.
We observe similar trends as in~\Cref{fig:acc_clust_cifar}: \sys shows comparable test accuracy for the nodes in the majority cluster and significantly higher test accuracy for nodes in the minority cluster.
\Cref{fig:acc_per_cluster_imagenette_20_4} (right) shows that after $ T = 800 $, \sys achieves 64.1\% test accuracy compared to 56.1\% test accuracy for \EL.
Finally, \Cref{fig:acc_clust_flickr} (in the appendix) shows that \sys also achieves a significant boost in model utility for the minority group in a 14:2 cluster configuration.

\begin{figure}[t!]
    \begin{subfigure}{0.99\linewidth} %
      \includegraphics[width=\linewidth]{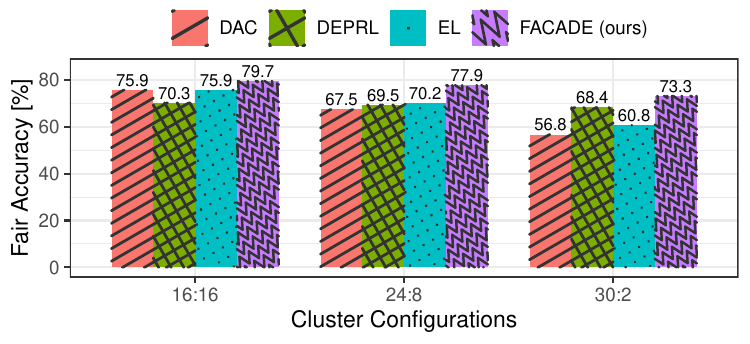}
      \caption{\cifar}\label{subfig:fairness_cifar10_barplot}
    \end{subfigure}\hfill
    \begin{subfigure}{0.99\linewidth} %
    	\includegraphics[width=\linewidth]{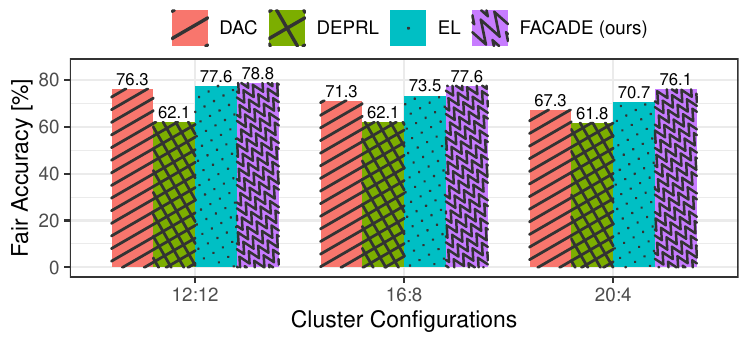}
    	\caption{\imagenette}\label{subfig:fairness_imagenette_barplot}
    \end{subfigure}\hfill
\caption{Highest observed fair accuracy for \cifar (top) and \imagenette (bottom), for varying cluster configurations and algorithms ($\uparrow$ is better).}
\label{fig:fairness_barplots}
\end{figure}

\subsection{Fair accuracy for varying cluster configurations}
\label{sec:exp_fair_accuracy}

We observe that the fairness metrics introduced in~\Cref{sec:bg_fairness} primarily focus on ensuring that the model’s predictions are independent of the different clusters in the network.
However, this overlooks the actual performance of the models, as a model that performs equally poorly across all groups could still receive a favorable fairness score.
For example, a random model achieving merely $10\%$ accuracy on a ten-class dataset ($\sim$ random guessing) would exhibit the highest possible fairness under these metrics simply because there would be no performance disparity between the groups.

\subsubsection{Fair accuracy metric}
\label{subsubsec:fair_acc_metric}
In order to demonstrate a more comprehensive evaluation of the performance of \sys by considering both fairness and accuracy, we introduce the notion of \emph{fair accuracy}.
Reflecting to the issue illustrated in \Cref{fig:acc_dpsgd_intro}, we recall that one of the primary motivations behind \sys is to reduce the performance gap between different clusters while maintaining high overall accuracy.
We introduce the \emph{fair accuracy} metric, which captures this by balancing the goal of achieving high overall accuracy while minimizing performance disparities between the clusters.
For $\lambda\in [0,1]$, the normalized \emph{fair accuracy}, denoted by $\operatorname{Acc}_{\operatorname{fair}}$, is defined as follows: %
\begin{align}
    &\operatorname{Acc}_{\operatorname{fair}}=\frac{\lambda \sum_{j\in [k]} \operatorname{Acc}_j}{k} + (1-\lambda)P, \label{eq:fair_acc_gen}
\end{align}
where $P=(1-(\max_{j\in [k]}\operatorname{Acc}_j - \min_{j\in [k]}\operatorname{Acc}_j))$ acts as a penalty term capturing the difference in the accuracy between the most and the least accurate clusters with $\operatorname{Acc}_j$ denoting the normalized average model accuracy for the $j^{th}$ cluster for every $j\in[k]$, and the hyperparameter $\lambda\in[0,1]$ assigns the context-specific weight to the average accuracy and the difference in accuracy between the clusters. 
In an ideal world, fair accuracy reaches its maximum value of 1 when there is no difference in the average model accuracies across the clusters.

\subsubsection{Fair accuracy of \sys and baselines}
We now present the performance of \sys and the baselines w.r.t. the fair accuracy metric given by \Cref{eq:fair_acc_gen}.
To compute this metric, we use $\lambda = \nicefrac{2}{3}$, as it slightly favors well-performing models while still giving significant weight to penalizing large discrepancies.
We use a similar experiment setup as in~\Cref{sec:exp_test_accuracy} and measure the fair accuracy periodically throughout the model training.

\Cref{fig:fairness_barplots} illustrates the highest obtained fair accuracy for each evaluated algorithm and cluster configuration, for both the \cifar and \imagenette datasets.
\Cref{subfig:fairness_cifar10_barplot} shows that \sys achieves the highest fair accuracy for all cluster configurations compared to baseline approaches.
For the 30:2 cluster configuration, \sys achieves 73.3\% fair accuracy, whereas the second-best baseline, \deprl, achieves 68.4\% fair accuracy.
In contrast, \EL shows a lower fair accuracy of 60.8\%.
This result underlines that \sys both achieves high model accuracy and minimizes the performance differences between groups.
The same trend holds for the \imagenette dataset, shown in~\Cref{subfig:fairness_imagenette_barplot}.
For the 20:4 cluster configuration, \sys achieves 76.1\% fair accuracy, compared to 70.7\% fair accuracy for \EL.
The fair accuracy for \flickr is provided in~\Cref{app:flickr_results} and is consistent with the findings for the \cifar and \imagenette datasets.
We provide additional fair accuracy plots in~\Cref{app:exp_fair_accuracy}, showing how the fair accuracy evolves throughout the training process.

\begin{figure}[t!]
	\begin{subfigure}{\columnwidth}
		\centering
		\includegraphics[width=.8\columnwidth]{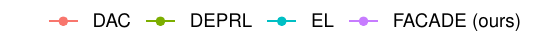}
	\end{subfigure}
    \begin{subfigure}{0.99\linewidth} %
      \includegraphics[width=\linewidth]{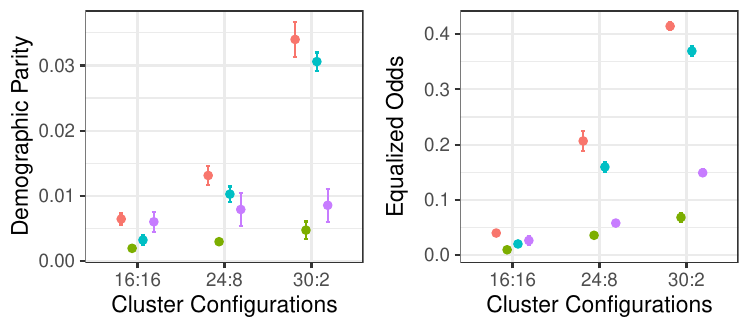}
      \caption{\cifar}\label{subfig:demo_par}
    \end{subfigure}\hfill
    \begin{subfigure}{\columnwidth}
    	\centering
    	\includegraphics[width=.8\columnwidth]{figures/fairness_legend.pdf}
    \end{subfigure}
    \begin{subfigure}{0.99\linewidth} %
    	\includegraphics[width=\linewidth]{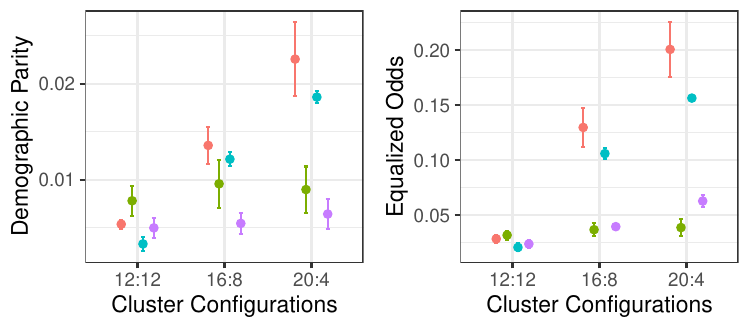}
    	\caption{\imagenette}\label{subfig:fairness_imagenette}
    \end{subfigure}\hfill
\caption{Boxplot of \acl{DP} (left, $\downarrow$ is better) and \acl{EO} (right, $\downarrow$ is better) obtained on \cifar and \imagenette, for varying cluster configurations and algorithms.}
\label{fig:fairness_plots}
\end{figure}

\subsection{Per-cluster fairness for varying cluster configurations}
\label{sec:exp_fairness}
We measure the \acf{DP} and \acf{EO} metrics on the final model of each experiment conducted in the previous section.
For this, we use~\Cref{eq:DP_overall} and~\Cref{eq:EO_overall} that were introduced in~\Cref{sec:bg_fairness}.
These results are visualized in~\Cref{fig:fairness_plots}, showing the \ac{DP} and \ac{EO} for different cluster configurations, algorithms, and for the \cifar and \imagenette datasets.

\Cref{fig:fairness_plots} shows that when cluster sizes are equal (\eg, 16:16 for \cifar and 12:12 for \imagenette), the \ac{DP} and \ac{EO} of all algorithms is comparable.
When the cluster sizes become more imbalanced, \sys exhibits lower \ac{DP} and \ac{EO} compared to \EL and \dac.
The only baseline that outperforms \sys is \deprl, which exhibits lower \ac{DP} and \ac{EO}.
However, regardless of cluster membership, the accuracy for all nodes with \deprl is lower, particularly for \imagenette (see~\Cref{fig:acc_clust_imagenette}), which can misleadingly be interpreted as having good fairness.
We notice that model heads in \deprl overfit significantly on the nodes' local data because it is never shared with other nodes.
Consequently, the algorithm cannot leverage the similar data distribution of other nodes and struggles to generalize on the test set.
This example motivates the fair accuracy metric we introduced in~\Cref{subsubsec:fair_acc_metric}.
Considering the low accuracy of \deprl, we conclude that \sys results in fairer treatment of minority clusters when cluster sizes are heavily imbalanced.

\subsection{Communication cost of \sys and baselines}
\label{sec:exp_communication_cost}
We now quantify the communication cost of \sys and baselines to reach a target accuracy.
We set this target accuracy to 63\% and 65\% for the \cifar and \imagenette datasets, respectively, which is the lowest accuracy by any baseline reached after $ T = 1200 $.
We exclude \deprl from this experiment due to its inferior performance compared to \sys and other baselines.
In contrast to the previous experiments, we consider the average accuracy of the entire network and not the per-cluster performance.

\begin{figure}[t!]
    \begin{subfigure}{0.99\linewidth} %
      \includegraphics[width=\linewidth]{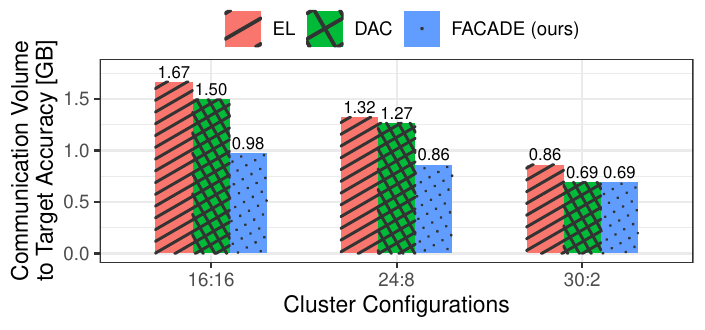}
      \caption{\cifar (target test accuracy: 63\%)}\label{fig:acc_bytes_cifar10}
    \end{subfigure}\hfill
    \begin{subfigure}{0.99\linewidth} %
    	\includegraphics[width=\linewidth]{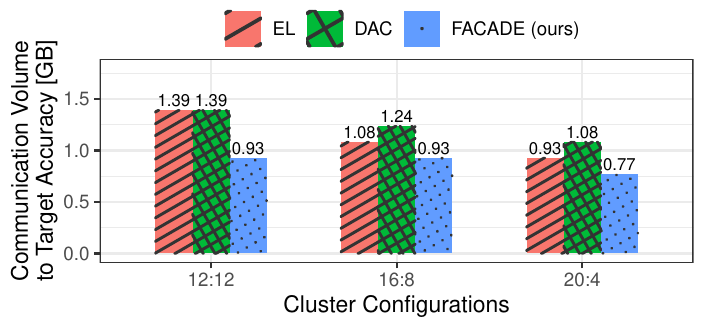}
    	\caption{\imagenette (target test accuracy: 65\%)}\label{fig:acc_bytes_imagenette}
    \end{subfigure}\hfill
\caption{Communication volume (in GB) required to reach a target test accuracy on the \cifar and \imagenette datasets, for different cluster configurations and algorithms ($\downarrow$ is better).}
\label{fig:acc_bytes}
\end{figure}

\Cref{fig:acc_bytes} illustrates the communication cost in GB required to reach the target accuracy for each cluster configuration and different algorithms for the \cifar and \imagenette datasets.
\Cref{fig:acc_bytes_cifar10} shows these results for \cifar.
In the 16:16 cluster configuration, \sys requires 41.3\% and 34.6\% less communication volume to reach the target accuracy compared to \EL and \dac, respectively.
This reduction is less pronounced for the 30:2 cluster configuration, with \sys requiring near-equal communication cost as \dac.
\Cref{fig:acc_bytes_imagenette} shows the communication volume required to reach the target accuracy for \imagenette.
In all cluster configurations, \sys requires less communication volume than \EL and \dac.
In the 16:16 cluster configuration, \sys requires 33.3\% less communication volume to reach the target accuracy than both \EL and \dac.
This reduction becomes 16.6\% and 28.5\% in the 20:4 cluster configuration compared to \EL and \dac, respectively.

From an algorithmic perspective, the communication cost of \sys per round is the same as that of \EL or \dpsgd.
Specifically, in each round, a node in \sys also sends only one model to each neighbor.
However, each model transfer includes an additional integer that indicates the model head index.
The overhead of this additional integer on the total communication volume is negligible.
Yet, \sys can reach the target accuracy faster than or equally fast as baselines.

\subsection{The effect of $ k $ on per-cluster test accuracy}
\label{sec:exp_estimate_k}
\sys introduces a hyperparameter $ k $, which defines the number of model heads each node maintains.
Ideally, $ k $ should match the number of unique features in the dataset, which can be difficult to estimate beforehand without sharing data.
We evaluate the sensitivity of \sys to variations in $ k $.
We create three clusters with 20, 10, and 2 nodes in a 32-node network and use the \cifar dataset.
Each cluster has images rotated by 0°, 90°, and 180°, respectively.
This experiment also demonstrates the capacity of \sys to handle more than two clusters.
We vary the number of model heads from one to five. %

\begin{figure}[t]
	\includegraphics[width=\linewidth]{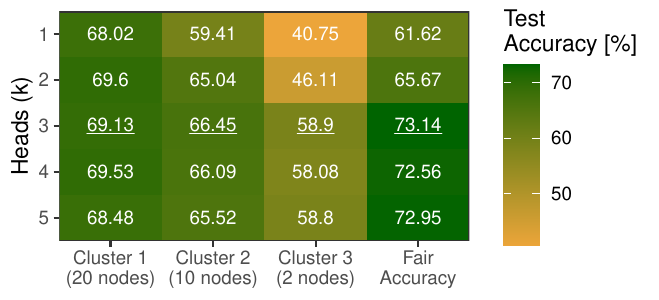}
	\caption{Highest attained test accuracy of \sys with a varying number of model heads for different clusters (left, $\uparrow$ is better). The average accuracies achieved by all nodes within each cluster are reported. We also report the fair accuracy in the right-most column.}
	\label{fig:acc_different_heads}
\end{figure}

\Cref{fig:acc_different_heads} shows the per-cluster test accuracy (columns) while varying $ k $ (rows).
We also show the fair accuracy in the right-most column.
When only one model head is used ($ k = 1 $), \sys is equivalent to \EL.
However, $ k = 1 $ shows poor test accuracy (40.75\%) for the minority cluster with only two nodes.
In the configuration with two model heads, two clusters tend to \emph{share} a head, while the remaining cluster has a dedicated model head that is specialized for its data distribution.
Using three model heads demonstrates the best performance for each cluster, except for the majority cluster.
This is expected as we augmented the data with three clusters of features.
Nevertheless, even when using four and five model heads, the attained model utility is remarkably close to those obtained with the optimal number of heads.
We observe that for $ k > 3 $, multiple heads tend to specialize for the same cluster, often the largest one.

This experiment highlights the robustness of our algorithm to variations of the hyperparameter $k$.
The system maintains a performance level close to the optimum, showcasing its resilience.
\Cref{fig:acc_different_heads} reveals that it is better to overestimate than to underestimate the value of $ k $.
However, in practical settings, there may be guidelines to estimate $ k $, \eg, the number of data sources used to obtain the training data.

\subsection{\sys cluster assignment}
\label{subsec:exp_settlement}
Finally, we analyze how nodes in \sys gravitate towards a particular cluster throughout the training process.
We use the same three-cluster setup as the experiment in~\Cref{sec:exp_estimate_k} and focus on the \cifar dataset.
We record for each node the cluster it belongs to and the model head index it selects during each communication round.

\begin{figure}[t]
	\centering
	\begin{subfigure}{\columnwidth}
		\centering
		\includegraphics[width=.65\columnwidth]{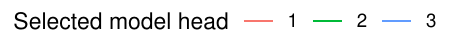}
	\end{subfigure}
	\begin{subfigure}{\columnwidth}
		\centering
		\includegraphics[width=\columnwidth]{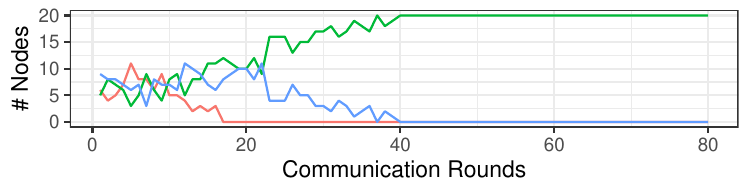}
		\caption{Cluster 1 (20 nodes)}
		\label{fig:settlement_cifar10_cluster0}
	\end{subfigure}
	\begin{subfigure}{\columnwidth}
		\centering
		\includegraphics[width=\columnwidth]{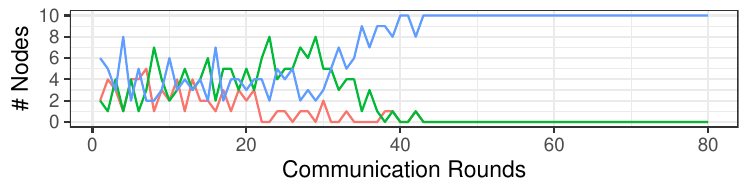}
		\caption{Cluster 2 (10 nodes)}
		\label{fig:settlement_cifar10_cluster1}
	\end{subfigure}
	\begin{subfigure}{\columnwidth}
		\centering
		\includegraphics[width=\columnwidth]{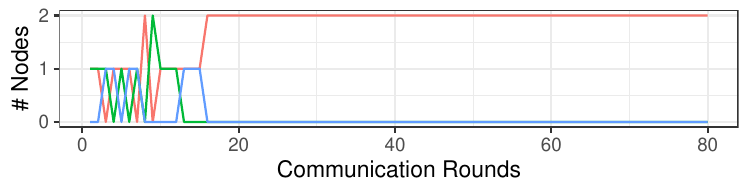}
		\caption{Cluster 3 (2 nodes)}
		\label{fig:settlement_cifar10_cluster2}
	\end{subfigure}%
	\caption{The distribution of model head selection by nodes in each of the three clusters during the first \num{80} communication round, using the \cifar dataset.}
	\label{fig:settling_evo_distr}
\end{figure}

\Cref{fig:settling_evo_distr} shows the distribution of model head selection by nodes in each cluster during the first \num{80} communication round.
At the start, we have a \emph{fuzzy} phase where nodes tend to explore different heads across rounds.
The nodes in the minority cluster converge quicker and pick the same model head just after \num{18} rounds.
After about \num{42} rounds, all nodes in the same cluster converge to the same head.
This shows that nodes in each cluster in \sys quickly converge to the same model head, compared to the total training duration (\num{1200} rounds).

During our experiments, we observed runs in which some nodes within the same cluster did not favor the same model head, or some model heads had not been selected at all.
We attribute this to variance in the obtained training losses during the early stages of the learning process.
We point out that similar behavior has been reported by related work on clustered federated learning~\cite{Ghosh_2020_IFCA}.
Nevertheless, we found such failures in cluster assignment to be a rare occurrence (\Cref{app:settlement_analysis}).

\section{Related Work}

\subsection {Personalized learning}
\label{sec:bg_personalized}
\sys can be seen as a form of personalized learning to realize per-cluster models with fair treatment for minority groups.
Personalization in the context of \ac{CML} involves training different models that are personalized for groups or individual nodes~\cite{kulkarni2020survey}.
In contrast, the objective of standard \ac{CML} algorithms such as FedAvg is to train a single, global model that meets the demands of all users.
Personalization can, for example, be used to deal with heterogeneous data distributions, where personalization can increase model utility across various subsets of the network compared to when using non-personalized approaches~\cite{ma2022state, tan2022towards, li2022federated}.

Personalization is a popular research avenue in \ac{FL}.
A prominent personalization approach involves each node maintaining a personal model, in addition to training a global one~\cite{huang2021personalized, li2021ditto, zhang2020fedfomo, mansour2020three}.

Similar to \sys, some other personalization techniques also involve nodes sharing only the \emph{core} of their model with the server, while the personalized \emph{head} is kept and learns to fit the local data~\cite{collins2021exploiting,caruana1997multitask,Ghosh_2020_IFCA,liang2020think}.
When the union of nodes' training data naturally forms clusters, some \FL techniques first partition these nodes accordingly before learning a separate model for each of them~\cite{sattler2020clustered, luo2021energy, dennis2021heterogeneity}.
A key difference between the work as mentioned above and \sys is that personalization in \FL is generally an easier problem to address than in a decentralized setting, as the server can collect global statistics and orchestrate the learning process.
Furthermore, in contrast to our work, none of the previously mentioned works focus on fairness.

IFCA~\cite{Ghosh_2020_IFCA} is similar to \sys in the sense that participants in IFCA also maintain multiple models and share the model exhibiting the lowest loss. However, there are two important differences between them: \emph{(i)} our work focuses on fairness, and \emph{(ii)} this work considers a decentralized setting whereas IFCA is a \ac{FL} algorithm which assumes a trusted central server.
More specifically, learning in decentralized networks can be more challenging than in centralized ones as there is no server that enables the collection of global statistics.

In \DL, early work on personalization incorporates a notion of similarity between nodes when constructing the communication topology~\cite{vanhaesebrouck2017decentralized, bellet2018personalized}.
These approaches assume the availability of a similarity matrix, which is often not the case in real-world scenarios.
As privacy is a major concern in \DL, the data needed to determine the similarity between nodes typically cannot be shared.
Other methods leverage a dynamic network to enhance personalization where neighbors are dynamically selected based on some performance metric throughout the training process~\cite{li2022learning,zec2022dac}.
\textsc{L2C}, for example, uses an attention mechanism to automatically assign mixing weights by comparing two nodes’ model updates~\cite{li2022learning}.
\sys, however, relies on unbiased, randomized communication, refreshing the topology each round.
Personalization has also been used to reduce the communication cost of \DL, \eg, \textsc{DisPFL} employs personalized sparse masks to customize sparse local models on the edge~\cite{dai2022dispfl}.
\deprl is a state-of-the-art personalized \ac{DL} algorithm that allows each node to optimize its model head locally while sharing the core model through periodic communication and aggregation steps~\cite{xiong2024deprl}. Unlike \sys, most existing personalization methods focus on personalization at the node level and do not leverage similarities when dealing with clustered data.
Additionally, these algorithms often focus on achieving high model utility and do not focus on fairness.

\subsection{Fairness in \ac{CML}}
The fairness-aware approaches in \ac{CML} usually aim to ensure equal model performance for all participants in the network.
The fairness objective in the work of Du et al.~\cite{du2021fairness} is to ensure that predictions made by the global model are fair regardless of variations in local data distributions of the clients
Similar to \sys, they use \ac{DP} as a fairness metric but operate in a centralized setting.
Chen et al.~\cite{chen2022fairness} propose a fairness-aware decentralized learning framework that uses a custom aggregation scheme considering the local learning objective and preferences.
FairFed~\cite{ezzeldin2023fairfed} is a \ac{FL} algorithm designed to enhance group fairness through a fairness-aware aggregation strategy.
To the best of our knowledge, we are the first to study fairness across different demographic groups in \ac{DL}.

\subsection{Other forms of data heterogeneity}
\sys is designed to achieve fairness in settings with feature heterogeneity.
In \ac{DL}, data heterogeneity may also arise from skewed label distributions or differences in the sizes of local datasets.
We included additional experiments with label skewness in~\Cref{app:label_heterogeneity}.
Dynamic topologies have been shown to allow better mixing of models in the presence of skewed label distributions~\cite{devos2024epidemic}.
Skewscout~\cite{hsieh2020non} reduces the communication cost in \ac{DL} in the presence of data heterogeneity.
In contrast, \sys tries to optimize both fairness and accuracy of the models.
We believe \sys is general enough to incorporate solutions like Skewscout to achieve fairness at lower communication costs.
However, communication cost is an orthogonal problem and \sys does not increase the communication compared to \ac{DL} algorithms such as \ac{EL}.

\section{Conclusion}
\label{sec:conclusion}
We introduced \sys, a novel \ac{DL} algorithm designed to address fairness issues in networks with feature heterogeneity.
By maintaining multiple model heads at each node, \sys personalizes models for different data clusters, ensuring fairness and high utility across different groups.
Our comprehensive evaluation with three state-of-the-art baselines and datasets demonstrates that \sys achieves high accuracy, particularly benefiting minority groups, without increasing communication costs.
\sys offers a robust solution for \ac{DL} in scenarios with feature heterogeneity, combining fairness and model accuracy.

\section*{Acknowledgement}
This work has been funded by the Swiss National Science Foundation, under the project ``FRIDAY: Frugal, Privacy-Aware and Practical Decentralized Learning'', SNSF proposal No. 10.001.796.

\bibliographystyle{IEEEtran}
\bibliography{IEEEabrv,main}

\begin{thebibliography}{10}
\providecommand{\url}[1]{#1}
\csname url@samestyle\endcsname
\providecommand{\newblock}{\relax}
\providecommand{\bibinfo}[2]{#2}
\providecommand{\BIBentrySTDinterwordspacing}{\spaceskip=0pt\relax}
\providecommand{\BIBentryALTinterwordstretchfactor}{4}
\providecommand{\BIBentryALTinterwordspacing}{\spaceskip=\fontdimen2\font plus
\BIBentryALTinterwordstretchfactor\fontdimen3\font minus
  \fontdimen4\font\relax}
\providecommand{\BIBforeignlanguage}[2]{{%
\expandafter\ifx\csname l@#1\endcsname\relax
\typeout{** WARNING: IEEEtran.bst: No hyphenation pattern has been}%
\typeout{** loaded for the language `#1'. Using the pattern for}%
\typeout{** the default language instead.}%
\else
\language=\csname l@#1\endcsname
\fi
#2}}
\providecommand{\BIBdecl}{\relax}
\BIBdecl

\bibitem{lian2017can}
X.~Lian, C.~Zhang, H.~Zhang, C.-J. Hsieh, W.~Zhang, and J.~Liu, ``Can
  decentralized algorithms outperform centralized algorithms? a case study for
  decentralized parallel stochastic gradient descent,'' \emph{Advances in
  neural information processing systems}, vol.~30, 2017.

\bibitem{ormandi2013gossip}
R.~Orm{\'a}ndi, I.~Heged{\H{u}}s, and M.~Jelasity, ``Gossip learning with
  linear models on fully distributed data,'' \emph{Concurrency and Computation:
  Practice and Experience}, vol.~25, no.~4, pp. 556--571, 2013.

\bibitem{devos2024epidemic}
M.~de~Vos, S.~Farhadkhani, R.~Guerraoui, A.-M. Kermarrec, R.~Pires, and
  R.~Sharma, ``Epidemic learning: Boosting decentralized learning with
  randomized communication,'' in \emph{NeurIPS}, 2023.

\bibitem{caldas2018leaf}
S.~Caldas, S.~M.~K. Duddu, P.~Wu, T.~Li, J.~Kone{\v{c}}n{\`y}, H.~B. McMahan,
  V.~Smith, and A.~Talwalkar, ``Leaf: A benchmark for federated settings,''
  \emph{arXiv preprint arXiv:1812.01097}, 2018.

\bibitem{rieke2020future}
N.~Rieke, J.~Hancox, W.~Li, F.~Milletari, H.~R. Roth, S.~Albarqouni, S.~Bakas,
  M.~N. Galtier, B.~A. Landman, K.~Maier-Hein \emph{et~al.}, ``The future of
  digital health with federated learning,'' \emph{NPJ digital medicine},
  vol.~3, no.~1, pp. 1--7, 2020.

\bibitem{nguyen2022federated}
D.~C. Nguyen, Q.-V. Pham, P.~N. Pathirana, M.~Ding, A.~Seneviratne, Z.~Lin,
  O.~Dobre, and W.-J. Hwang, ``Federated learning for smart healthcare: A
  survey,'' \emph{ACM Computing Surveys (Csur)}, vol.~55, no.~3, pp. 1--37,
  2022.

\bibitem{lecun1998gradient}
Y.~LeCun, L.~Bottou, Y.~Bengio, and P.~Haffner, ``Gradient-based learning
  applied to document recognition,'' \emph{Proceedings of the IEEE}, vol.~86,
  no.~11, pp. 2278--2324, 1998.

\bibitem{krizhevsky2009learning}
A.~Krizhevsky, G.~Hinton \emph{et~al.}, ``Learning multiple layers of features
  from tiny images,'' 2009.

\bibitem{pessach2022review}
D.~Pessach and E.~Shmueli, ``A review on fairness in machine learning,''
  \emph{ACM Computing Surveys (CSUR)}, vol.~55, no.~3, pp. 1--44, 2022.

\bibitem{berk2021fairness}
R.~Berk, H.~Heidari, S.~Jabbari, M.~Kearns, and A.~Roth, ``Fairness in criminal
  justice risk assessments: The state of the art,'' \emph{Sociological Methods
  \& Research}, vol.~50, no.~1, pp. 3--44, 2021.

\bibitem{xiong2024deprl}
G.~Xiong, G.~Yan, S.~Wang, and J.~Li, ``Deprl: Achieving linear convergence
  speedup in personalized decentralized learning with shared representations,''
  in \emph{Proceedings of the AAAI Conference on Artificial Intelligence},
  vol.~38, no.~14, 2024, pp. 16\,103--16\,111.

\bibitem{zec2022dac}
E.~L. Zec, E.~Ekblom, M.~Willbo, O.~Mogren, and S.~Girdzijauskas,
  ``Decentralized adaptive clustering of deep nets is beneficial for client
  collaboration,'' in \emph{International Workshop on Trustworthy Federated
  Learning}.\hskip 1em plus 0.5em minus 0.4em\relax Springer, 2022, pp. 59--71.

\bibitem{soykan2022survey}
E.~U. Soykan, L.~Kara{\c{c}}ay, F.~Karako{\c{c}}, and E.~Tomur, ``A survey and
  guideline on privacy enhancing technologies for collaborative machine
  learning,'' \emph{IEEE Access}, vol.~10, 2022.

\bibitem{pasquini2023security}
D.~Pasquini, M.~Raynal, and C.~Troncoso, ``On the (in) security of peer-to-peer
  decentralized machine learning,'' in \emph{2023 IEEE Symposium on Security
  and Privacy (SP)}, 2023, pp. 418--436.

\bibitem{nedic2016stochastic}
A.~Nedi{\'c} and A.~Olshevsky, ``Stochastic gradient-push for strongly convex
  functions on time-varying directed graphs,'' \emph{IEEE Transactions on
  Automatic Control}, vol.~61, no.~12, 2016.

\bibitem{assran2019stochastic}
M.~Assran, N.~Loizou, N.~Ballas, and M.~Rabbat, ``Stochastic gradient push for
  distributed deep learning,'' in \emph{ICML}, 2019.

\bibitem{lu2023privacy}
Y.~Lu, Z.~Yu, and N.~Suri, ``Privacy-preserving decentralized federated
  learning over time-varying communication graph,'' \emph{ACM Transactions on
  Privacy and Security}, vol.~26, no.~3, pp. 1--39, 2023.

\bibitem{mcmahan2017fedavg}
B.~McMahan, E.~Moore, D.~Ramage, S.~Hampson, and B.~A. y~Arcas,
  ``Communication-efficient learning of deep networks from decentralized
  data,'' in \emph{Artificial intelligence and statistics}.\hskip 1em plus
  0.5em minus 0.4em\relax PMLR, 2017, pp. 1273--1282.

\bibitem{mehrabi2021survey}
N.~Mehrabi, F.~Morstatter, N.~Saxena, K.~Lerman, and A.~Galstyan, ``A survey on
  bias and fairness in machine learning,'' \emph{ACM computing surveys (CSUR)},
  vol.~54, no.~6, pp. 1--35, 2021.

\bibitem{verma2018fairness}
S.~Verma and J.~Rubin, ``Fairness definitions explained,'' in \emph{Proceedings
  of the international workshop on software fairness}, 2018, pp. 1--7.

\bibitem{hanna2009measuring}
R.~Hanna and L.~Linden, ``Measuring discrimination in education,'' National
  Bureau of Economic Research, Tech. Rep., 2009.

\bibitem{makhlouf2021applicability}
K.~Makhlouf, S.~Zhioua, and C.~Palamidessi, ``On the applicability of machine
  learning fairness notions,'' \emph{ACM SIGKDD Explorations Newsletter},
  vol.~23, no.~1, pp. 14--23, 2021.

\bibitem{dwork2012fairness}
C.~Dwork, M.~Hardt, T.~Pitassi, O.~Reingold, and R.~Zemel, ``Fairness through
  awareness,'' in \emph{Proceedings of the 3rd innovations in theoretical
  computer science conference}, 2012, pp. 214--226.

\bibitem{hardt2016equality}
M.~Hardt, E.~Price, and N.~Srebro, ``Equality of opportunity in supervised
  learning,'' \emph{Advances in neural information processing systems},
  vol.~29, 2016.

\bibitem{ezzeldin2023fairfed}
Y.~H. Ezzeldin, S.~Yan, C.~He, E.~Ferrara, and A.~S. Avestimehr, ``Fairfed:
  Enabling group fairness in federated learning,'' in \emph{Proceedings of the
  AAAI conference on artificial intelligence}, vol.~37, no.~6, 2023, pp.
  7494--7502.

\bibitem{galli2023advancing}
F.~Galli, K.~Jung, S.~Biswas, C.~Palamidessi, and T.~Cucinotta, ``Advancing
  personalized federated learning: Group privacy, fairness, and beyond,''
  \emph{SN Computer Science}, vol.~4, no.~6, p. 831, 2023.

\bibitem{denis2024fairness}
C.~Denis, R.~Elie, M.~Hebiri, and F.~Hu, ``Fairness guarantees in multi-class
  classification with demographic parity,'' \emph{Journal of Machine Learning
  Research}, vol.~25, no. 130, pp. 1--46, 2024.

\bibitem{rouzot2022learning}
J.~Rouzot, J.~Ferry, and M.~Huguet, ``Learning optimal fair scoring systems for
  multi-class classification,'' in \emph{2022 IEEE 34th International
  Conference on Tools with Artificial Intelligence (ICTAI)}.\hskip 1em plus
  0.5em minus 0.4em\relax Los Alamitos, CA, USA: IEEE Computer Society, nov
  2022, pp. 197--204.

\bibitem{beltran2023decentralized}
E.~T.~M. Beltr{\'a}n, M.~Q. P{\'e}rez, P.~M.~S. S{\'a}nchez, S.~L. Bernal,
  G.~Bovet, M.~G. P{\'e}rez, G.~M. P{\'e}rez, and A.~H. Celdr{\'a}n,
  ``Decentralized federated learning: Fundamentals, state of the art,
  frameworks, trends, and challenges,'' \emph{IEEE Communications Surveys \&
  Tutorials}, 2023.

\bibitem{shiranthika2023decentralized}
C.~Shiranthika, P.~Saeedi, and I.~V. Baji{\'c}, ``Decentralized learning in
  healthcare: a review of emerging techniques,'' \emph{IEEE Access}, vol.~11,
  2023.

\bibitem{douceur2002sybil}
J.~R. Douceur, ``The sybil attack,'' in \emph{International workshop on
  peer-to-peer systems}.\hskip 1em plus 0.5em minus 0.4em\relax Springer, 2002.

\bibitem{antonov2023securecyclon}
A.~Antonov and S.~Voulgaris, ``Securecyclon: Dependable peer sampling,'' in
  \emph{2023 IEEE 43rd International Conference on Distributed Computing
  Systems (ICDCS)}.\hskip 1em plus 0.5em minus 0.4em\relax IEEE, 2023, pp.
  1--12.

\bibitem{guerraoui2024peerswap}
R.~Guerraoui, A.-M. Kermarrec, A.~Kucherenko, R.~Pinot, and M.~de~Vos,
  ``Peerswap: A peer-sampler with randomness guarantees,'' in \emph{Proceedings
  of the 43rd International Symposium on Reliable Distributed Systems (SRDS
  2024)}, 2024.

\bibitem{Ghosh_2020_IFCA}
A.~Ghosh, J.~Chung, D.~Yin, and K.~Ramchandran, ``An efficient framework for
  clustered federated learning,'' in \emph{Proceedings of the 34th
  International Conference on Neural Information Processing Systems}, ser. NIPS
  '20.\hskip 1em plus 0.5em minus 0.4em\relax Red Hook, NY, USA: Curran
  Associates Inc., 2020.

\bibitem{macqueen1967some}
J.~MacQueen \emph{et~al.}, ``Some methods for classification and analysis of
  multivariate observations,'' in \emph{Proceedings of the fifth Berkeley
  symposium on mathematical statistics and probability}, vol.~1, no.~14.\hskip
  1em plus 0.5em minus 0.4em\relax Oakland, CA, USA, 1967, pp. 281--297.

\bibitem{reynolds2009gaussian}
D.~A. Reynolds \emph{et~al.}, ``Gaussian mixture models.'' \emph{Encyclopedia
  of biometrics}, vol. 741, no. 659-663, 2009.

\bibitem{cyffers2022muffliato}
E.~Cyffers, M.~Even, A.~Bellet, and L.~Massouli{\'e}, ``Muffliato: Peer-to-peer
  privacy amplification for decentralized optimization and averaging,''
  \emph{Advances in Neural Information Processing Systems}, vol.~35, pp.
  15\,889--15\,902, 2022.

\bibitem{biswas2024beyond}
S.~Biswas, M.~Even, A.-M. Kermarrec, L.~Massoulie, R.~Pires, R.~Sharma, and
  M.~de~Vos, ``Noiseless privacy-preserving decentralized learning,''
  \emph{arXiv preprint arXiv:2404.09536}, 2024.

\bibitem{biswas2024lowcost}
S.~Biswas, D.~Frey, R.~Gaudel, A.-M. Kermarrec, D.~Ler{\'e}v{\'e}rend,
  R.~Pires, R.~Sharma, and F.~Ta{\"\i}ani, ``Low-cost privacy-aware
  decentralized learning,'' \emph{arXiv preprint arXiv:2403.11795}, 2024.

\bibitem{dhasade2023decentralized}
A.~Dhasade, A.-M. Kermarrec, R.~Pires, R.~Sharma, and M.~Vujasinovic,
  ``Decentralized learning made easy with decentralizepy,'' in
  \emph{Proceedings of the 3rd Workshop on Machine Learning and Systems}, 2023,
  pp. 34--41.

\bibitem{hsieh2020non}
K.~Hsieh, A.~Phanishayee, O.~Mutlu, and P.~Gibbons, ``The non-iid data quagmire
  of decentralized machine learning,'' in \emph{International Conference on
  Machine Learning}.\hskip 1em plus 0.5em minus 0.4em\relax PMLR, 2020, pp.
  4387--4398.

\bibitem{imagenette}
\BIBentryALTinterwordspacing
J.~Howard, ``Imagenette.'' [Online]. Available:
  \url{https://github.com/fastai/imagenette/}
\BIBentrySTDinterwordspacing

\bibitem{deng2009imagenet}
J.~Deng, W.~Dong, R.~Socher, L.-J. Li, K.~Li, and L.~Fei-Fei, ``Imagenet: A
  large-scale hierarchical image database,'' in \emph{2009 IEEE conference on
  computer vision and pattern recognition}.\hskip 1em plus 0.5em minus
  0.4em\relax Ieee, 2009, pp. 248--255.

\bibitem{onoszko2021decentralized}
N.~Onoszko, G.~Karlsson, O.~Mogren, and E.~L. Zec, ``Decentralized federated
  learning of deep neural networks on non-iid data,'' \emph{arXiv preprint
  arXiv:2107.08517}, 2021.

\bibitem{chung2022federated}
J.~Chung, K.~Lee, and K.~Ramchandran, ``Federated unsupervised clustering with
  generative models,'' in \emph{AAAI 2022 international workshop on trustable,
  verifiable and auditable federated learning}, vol.~4, 2022.

\bibitem{he2016deep}
K.~He, X.~Zhang, S.~Ren, and J.~Sun, ``Deep residual learning for image
  recognition,'' in \emph{Proceedings of the IEEE conference on computer vision
  and pattern recognition}, 2016, pp. 770--778.

\bibitem{patarasuk2009bandwidth}
P.~Patarasuk and X.~Yuan, ``Bandwidth optimal all-reduce algorithms for
  clusters of workstations,'' \emph{Journal of Parallel and Distributed
  Computing}, vol.~69, no.~2, pp. 117--124, 2009.

\bibitem{kulkarni2020survey}
V.~Kulkarni, M.~Kulkarni, and A.~Pant, ``Survey of personalization techniques
  for federated learning,'' in \emph{2020 fourth world conference on smart
  trends in systems, security and sustainability (WorldS4)}.\hskip 1em plus
  0.5em minus 0.4em\relax IEEE, 2020, pp. 794--797.

\bibitem{ma2022state}
X.~Ma, J.~Zhu, Z.~Lin, S.~Chen, and Y.~Qin, ``A state-of-the-art survey on
  solving non-iid data in federated learning,'' \emph{Future Generation
  Computer Systems}, vol. 135, pp. 244--258, 2022.

\bibitem{tan2022towards}
A.~Z. Tan, H.~Yu, L.~Cui, and Q.~Yang, ``Towards personalized federated
  learning,'' \emph{IEEE Transactions on Neural Networks and Learning Systems},
  2022.

\bibitem{li2022federated}
Q.~Li, Y.~Diao, Q.~Chen, and B.~He, ``Federated learning on non-iid data silos:
  An experimental study,'' in \emph{2022 IEEE 38th international conference on
  data engineering (ICDE)}.\hskip 1em plus 0.5em minus 0.4em\relax IEEE, 2022,
  pp. 965--978.

\bibitem{huang2021personalized}
Y.~Huang, L.~Chu, Z.~Zhou, L.~Wang, J.~Liu, J.~Pei, and Y.~Zhang,
  ``Personalized cross-silo federated learning on non-iid data,'' in
  \emph{Proceedings of the AAAI conference on artificial intelligence},
  vol.~35, no.~9, 2021, pp. 7865--7873.

\bibitem{li2021ditto}
T.~Li, S.~Hu, A.~Beirami, and V.~Smith, ``Ditto: Fair and robust federated
  learning through personalization,'' in \emph{International conference on
  machine learning}.\hskip 1em plus 0.5em minus 0.4em\relax PMLR, 2021, pp.
  6357--6368.

\bibitem{zhang2020fedfomo}
M.~Zhang, K.~Sapra, S.~Fidler, S.~Yeung, and J.~M. Alvarez, ``Personalized
  federated learning with first order model optimization,'' \emph{arXiv
  preprint arXiv:2012.08565}, 2020.

\bibitem{mansour2020three}
Y.~Mansour, M.~Mohri, J.~Ro, and A.~T. Suresh, ``Three approaches for
  personalization with applications to federated learning,'' \emph{arXiv
  preprint arXiv:2002.10619}, 2020.

\bibitem{collins2021exploiting}
L.~Collins, H.~Hassani, A.~Mokhtari, and S.~Shakkottai, ``Exploiting shared
  representations for personalized federated learning,'' in \emph{International
  conference on machine learning}.\hskip 1em plus 0.5em minus 0.4em\relax PMLR,
  2021, pp. 2089--2099.

\bibitem{caruana1997multitask}
R.~Caruana, ``Multitask learning,'' \emph{Machine learning}, vol.~28, pp.
  41--75, 1997.

\bibitem{liang2020think}
P.~P. Liang, T.~Liu, L.~Ziyin, N.~B. Allen, R.~P. Auerbach, D.~Brent,
  R.~Salakhutdinov, and L.-P. Morency, ``Think locally, act globally: Federated
  learning with local and global representations,'' \emph{arXiv preprint
  arXiv:2001.01523}, 2020.

\bibitem{sattler2020clustered}
F.~Sattler, K.-R. M{\"u}ller, and W.~Samek, ``Clustered federated learning:
  Model-agnostic distributed multitask optimization under privacy
  constraints,'' \emph{IEEE transactions on neural networks and learning
  systems}, vol.~32, no.~8, pp. 3710--3722, 2020.

\bibitem{luo2021energy}
Y.~Luo, X.~Liu, and J.~Xiu, ``Energy-efficient clustering to address data
  heterogeneity in federated learning,'' in \emph{ICC 2021-IEEE International
  Conference on Communications}.\hskip 1em plus 0.5em minus 0.4em\relax IEEE,
  2021, pp. 1--6.

\bibitem{dennis2021heterogeneity}
D.~K. Dennis, T.~Li, and V.~Smith, ``Heterogeneity for the win: One-shot
  federated clustering,'' in \emph{International Conference on Machine
  Learning}.\hskip 1em plus 0.5em minus 0.4em\relax PMLR, 2021, pp. 2611--2620.

\bibitem{vanhaesebrouck2017decentralized}
P.~Vanhaesebrouck, A.~Bellet, and M.~Tommasi, ``Decentralized collaborative
  learning of personalized models over networks,'' in \emph{Artificial
  Intelligence and Statistics}.\hskip 1em plus 0.5em minus 0.4em\relax PMLR,
  2017, pp. 509--517.

\bibitem{bellet2018personalized}
A.~Bellet, R.~Guerraoui, M.~Taziki, and M.~Tommasi, ``Personalized and private
  peer-to-peer machine learning,'' in \emph{International conference on
  artificial intelligence and statistics}.\hskip 1em plus 0.5em minus
  0.4em\relax PMLR, 2018, pp. 473--481.

\bibitem{li2022learning}
S.~Li, T.~Zhou, X.~Tian, and D.~Tao, ``Learning to collaborate in decentralized
  learning of personalized models,'' in \emph{Proceedings of the IEEE/CVF
  Conference on Computer Vision and Pattern Recognition}, 2022, pp. 9766--9775.

\bibitem{dai2022dispfl}
R.~Dai, L.~Shen, F.~He, X.~Tian, and D.~Tao, ``Dispfl: Towards
  communication-efficient personalized federated learning via decentralized
  sparse training,'' in \emph{International conference on machine
  learning}.\hskip 1em plus 0.5em minus 0.4em\relax PMLR, 2022, pp. 4587--4604.

\bibitem{du2021fairness}
W.~Du, D.~Xu, X.~Wu, and H.~Tong, ``Fairness-aware agnostic federated
  learning,'' in \emph{Proceedings of the 2021 SIAM International Conference on
  Data Mining (SDM)}.\hskip 1em plus 0.5em minus 0.4em\relax SIAM, 2021, pp.
  181--189.

\bibitem{chen2022fairness}
Z.~Chen, W.~Liao, P.~Tian, Q.~Wang, and W.~Yu, ``A fairness-aware peer-to-peer
  decentralized learning framework with heterogeneous devices,'' \emph{Future
  Internet}, vol.~14, no.~5, p. 138, 2022.

\bibitem{lin2022towards}
W.~Lin, B.~Li, and C.~Wang, ``Towards private learning on decentralized graphs
  with local differential privacy,'' \emph{IEEE Transactions on Information
  Forensics and Security}, vol.~17, pp. 2936--2946, 2022.

\bibitem{mansouri2023sok}
M.~Mansouri, M.~{\"O}nen, W.~B. Jaballah, and M.~Conti, ``Sok: Secure
  aggregation based on cryptographic schemes for federated learning,''
  \emph{Proceedings on Privacy Enhancing Technologies}, 2023.

\end{thebibliography}

\clearpage
\appendices
\label{sec:appendix}

\section{\dpsgd Pseudocode}
\label[appendix]{app:dpsgd}
We show in~\Cref{alg:dpsgd} the standard \dpsgd procedure from the perspective of node $ i $.

\begin{algorithm2e}[h]
    \DontPrintSemicolon
    \caption{The \dpsgd procedure, from the perspective of node $i$.}
    \label{alg:dpsgd}
    Initialize $\theta_i^{(0)}$\;
    \For{$t = 0, \dots, T-1$}{
    $\tilde \theta_i^{(t,0)}\leftarrow\theta_i^{(t)}$\;
        \For{$h = 0, \dots, H-1$}{\label{line:local_update_start}
            $\xi_i \leftarrow$ mini-batch sampled from $D_i$\;
            $\tilde\theta_i^{(t,h+1)} \leftarrow \tilde\theta_i^{(t,h)} - \eta \nabla f_i(\tilde\theta_i^{(t,h)}, \xi_i)$\;
        }\label{line:local_update_end}
        \textbf{Send} $\tilde\theta_i^{(t,H)}$ to the neighbors in topology $ \cG $ \label{line:share}\;
       \textbf{Receive} $\tilde\theta_j^{(t,H)}$ from each neighbor $j$ in $ \cG $ \label{line:receive}\;
         \textbf{Aggregate} the received models to produce $\theta_i^{(t+1)}$\label{line:aggregate}\;
    }
    \KwRet $\theta_i^{(T)}$\;
\end{algorithm2e}

\section{Proofs}\label[appendix]{app:proofs}
\paragraph*{Proof of \Cref{th:local_convergence}}  
     From the perspective of a fixed node $N_i$, we can draw the same line of reasoning as in Theorem 2 and Lemma 3 of \cite{Ghosh_2020_IFCA}. \qed

\paragraph*{Proof of \Cref{th:global_convergence}} 

 Fix any cluster $j\in [k]$ and a round $t>0$. Firstly, we observe that
 \begin{align}
     &\norm{\theta^{[j](t+1)}-\theta^{*}_{[j]}}=\norm{\left(\frac{1}{|S^{(t)}_{[j]}|}\sum_{i\in S_{[j]}^{(t)}}\bar\theta_i^{[j](t+1)}\right)-\theta^{*}_{[j]}}\nonumber\\
     &=\norm{\frac{1}{|S^{(t)}_{[j]}|}\sum_{i\in S_{[j]}^{(t)}}\left(\bar\theta_i^{[j](t+1)}-\theta^{*}_{[j]}\right)}\nonumber\\
     &\leq \frac{1}{|S^{(t)}_{[j]}|} \sum_{i\in S_{[j]}^{(t)}}\norm{\bar\theta_i^{[j](t+1)}-\theta^{*}_{[j]}}\text{ [triangle inequality]}\label{eq:global_conv_1}
 \end{align}
From \eqref{eq:global_conv_1}, we conclude that for any $\zeta>0$, we have:
\begin{align}
         &\mathbb{P}\left[\norm{\bar\theta^{[j](t+1)}-\theta^{*}_{[j]}}\leq \zeta\right]\nonumber\\
         &\geq \mathbb{P}\left[ \frac{1}{|S^{(t)}_{[j]}|} \sum_{i\in S_{[j]}^{(t)}}\norm{\bar\theta_i^{[j](t+1)}-\theta^{*}_{[j]}}\leq \zeta\right]\nonumber\\
         &= \mathbb{P}\left[\sum_{i\in S_{[j]}^{(t)}}\norm{\bar\theta_i^{[j](t+1)}-\theta^{*}_{[j]}}\leq |S^{(t)}_{[j]}|\zeta\right]\nonumber\\
         &\geq \mathbb{P}\left[\bigcap_{i\in S_{[j]}^{(t)}}\norm{\bar\theta_i^{[j](t+1)}-\theta^{*}_{[j]}}\leq\zeta\right]\nonumber\\
         &= \prod_{i\in S_{[j]}^{(t)}}\mathbb{P}\left[\norm{\bar\theta_i^{[j](t+1)}-\theta^{*}_{[j]}}\leq\zeta\right]\label{eq:global_conv_2}\\
         &[\because\,\text{$\bar\theta_1^{[j](t+1)},\ldots,\bar\theta_n^{[j](t+1)}$\, %
         are pairwise-independent}]\nonumber
\end{align}
Setting $\zeta=(1-\frac{p\lambda}{8L}) \norm{\theta_i^{[j](t)}-\theta^{*}_{[j]}}+\epsilon_0$ and using \Cref{th:local_convergence}, we conclude the proof. \qed

\paragraph*{Proof of \Cref{th:final_convergence}} 
Using \Cref{th:global_convergence}, the proof follows the identical line of reasoning as that of Corollary 2 of \cite{Ghosh_2020_IFCA}. \qed

\section{Experimental setup}
\label[appendix]{app:expt_setup}
\Cref{table:experiment_datasets} summarizes the used dataset and adopted learning parameters for each dataset.

\begin{table*}[t!]
	\small
    \caption{Summary of datasets and parameters used to evaluate \sys and \ac{DL} baselines.}
	\centering
	\begin{tabular}{c|c|c|c|c|c|c|c}
		\toprule
		\multirow{2}{*}{\textsc{Dataset}} & \multirow{2}{*}{\textsc{Nodes}} & \multirow{2}{*}{\textsc{Model}} & \multirow{2}{*}{\textsc{Model Params.}} & \multicolumn{4}{c}{\textsc{Learning Rates}} \\
		\cmidrule(lr){5-8}
		& & & & \EL & \dac & \deprl & \sys \\ 
		\midrule
		\cifar~\cite{krizhevsky2009learning} & 32 & CNN (LeNet~\cite{lecun1998gradient})  & 120k & $ \eta = 0.05$ & $ \eta = 0.005 $ & $ \eta = 0.01 $ & $ \eta = 0.01 $ \\ 
		\imagenette~\cite{imagenette} & 24 & CNN (LeNet~\cite{lecun1998gradient})  & 250k & $ \eta = 0.001 $ & $ \eta = 0.001 $ & $ \eta = 0.0005 $ & $ \eta = 0.0003 $ \\ 
		\flickr~\cite{hsieh2020non} & 16 & ResNet8~\cite{he2016deep} & 310k & $ \eta = 0.1 $ & $ \eta = 0.3 $ & $ \eta = 0.1 $ & $ \eta = 0.3 $ \\ 
		\bottomrule
	\end{tabular}
	\label{table:experiment_datasets}
\end{table*}

\begin{figure*}[t!]
	\centering
	\begin{subfigure}{\columnwidth}
		\centering
		\includegraphics[width=.8\columnwidth]{figures/acc_per_cluster_legend.pdf}
	\end{subfigure}
	\begin{subfigure}{.33\textwidth}
		\centering
		\includegraphics[width=\textwidth]{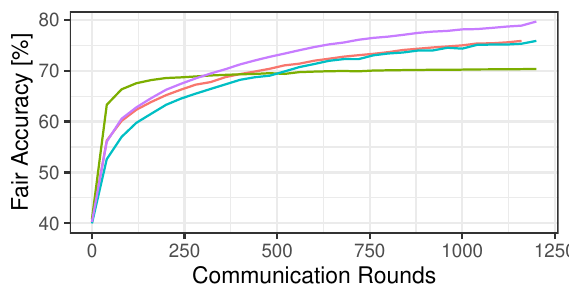}
		\caption{16:16 cluster configuration}
		\label{fig:fair_acc_cifar10_16_16}
	\end{subfigure}%
	\begin{subfigure}{.33\textwidth}
		\centering
		\includegraphics[width=\textwidth]{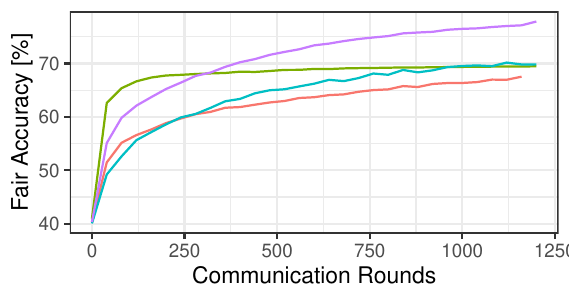}
		\caption{24:8 cluster configuration}
		\label{fig:fair_acc_cifar10_24_8}
	\end{subfigure}%
	\begin{subfigure}{.33\textwidth}
		\centering
		\includegraphics[width=\textwidth]{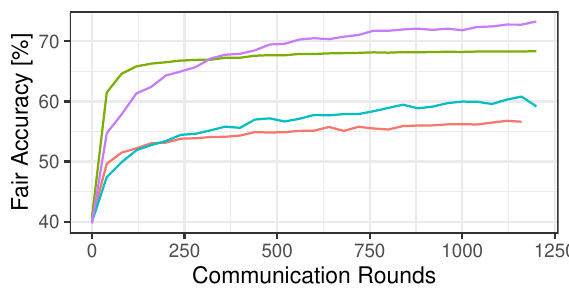}
		\caption{30:2 cluster configuration}
		\label{fig:fair_acc_cifar10_30_2}
	\end{subfigure}%
	\caption{Fair Accuracy ($\uparrow$ is better) obtained on \cifar.}
	\label{fig:fair_acc_cifar10}
\end{figure*}

\begin{figure*}[t]
	\centering
	\begin{subfigure}{\columnwidth}
		\centering
		\includegraphics[width=.8\columnwidth]{figures/acc_per_cluster_legend.pdf}
	\end{subfigure}
	\begin{subfigure}{.33\textwidth}
		\centering
		\includegraphics[width=\textwidth]{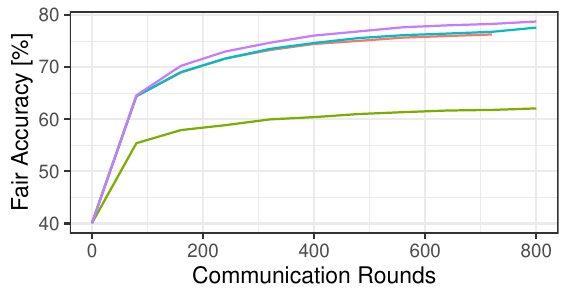}
		\caption{12:12 cluster configuration}
		\label{fig:fair_acc_imagenette_12_12}
	\end{subfigure}%
	\begin{subfigure}{.33\textwidth}
		\centering
		\includegraphics[width=\textwidth]{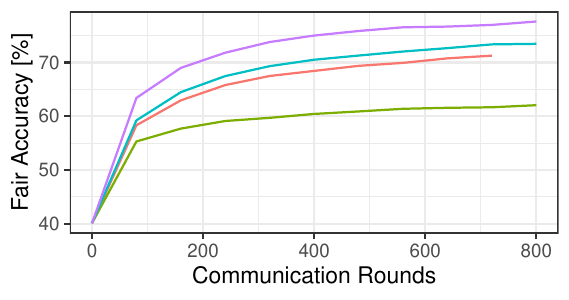}
		\caption{16:8 cluster configuration}
		\label{fig:fair_acc_imagenette_16_8}
	\end{subfigure}%
	\begin{subfigure}{.33\textwidth}
		\centering
		\includegraphics[width=\textwidth]{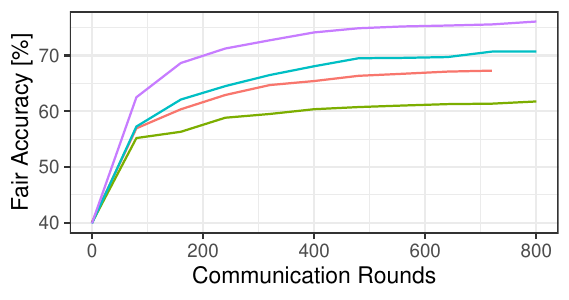}
		\caption{20:4 cluster configuration}
		\label{fig:fair_acc_imagenette_20_4}
	\end{subfigure}%
	\caption{Fair Accuracy ($\uparrow$ is better) obtained on \imagenette.}
	\label{fig:fair_acc_imagenette}
\end{figure*}

\begin{figure*}[t]
	\centering
	\begin{subfigure}{\columnwidth}
		\centering
		\includegraphics[width=.8\columnwidth]{figures/acc_per_cluster_legend.pdf}
	\end{subfigure}
	\begin{subfigure}{.33\textwidth}
		\centering
		\includegraphics[width=\textwidth]{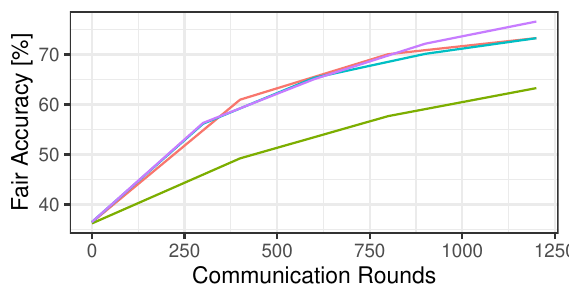}
		\caption{8:8 cluster configuration}
		\label{fig:fair_acc_flickr_8_8}
	\end{subfigure}%
	\begin{subfigure}{.33\textwidth}
		\centering
		\includegraphics[width=\textwidth]{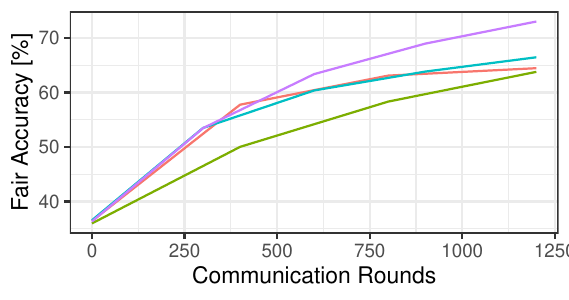}
		\caption{14:2 cluster configuration}
		\label{fig:fair_acc_flickr_14_2}
	\end{subfigure}%
	\caption{Fair Accuracy ($\uparrow$ is better) obtained on \flickr.}
	\label{fig:fair_acc_flickr}
\end{figure*}

\section{Additional fair accuracy plots}
\label[appendix]{app:exp_fair_accuracy}
\Cref{fig:fair_acc_cifar10}, \Cref{fig:fair_acc_imagenette} and \Cref{fig:fair_acc_flickr} provide the evolution of fair accuracy for varying cluster configurations and the \cifar, \imagenette, and \flickr datasets, respectively.
To compute the fair accuracy, we use $\alpha = \nicefrac{2}{3}$.
We observe that for all datasets and cluster configurations, \sys achieves the highest fair accuracy.

\begin{figure}[t]
	\centering
	\begin{subfigure}{\columnwidth}
		\centering
		\includegraphics[width=.8\columnwidth]{figures/acc_per_cluster_legend.pdf}
	\end{subfigure}
	\begin{subfigure}{\columnwidth}
		\centering
		\includegraphics[width=\columnwidth]{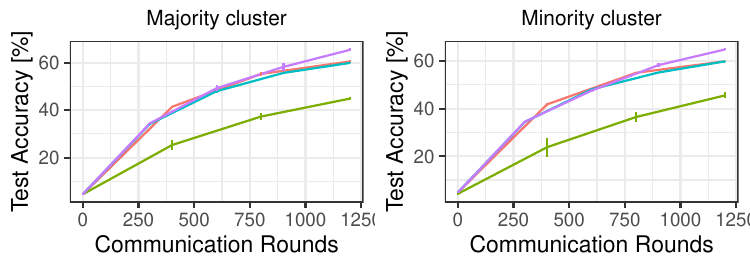}
		\caption{8:8 cluster configuration}
		\label{fig:acc_per_cluster_flickr_8_8}
	\end{subfigure}
	\begin{subfigure}{\columnwidth}
		\centering
		\includegraphics[width=\columnwidth]{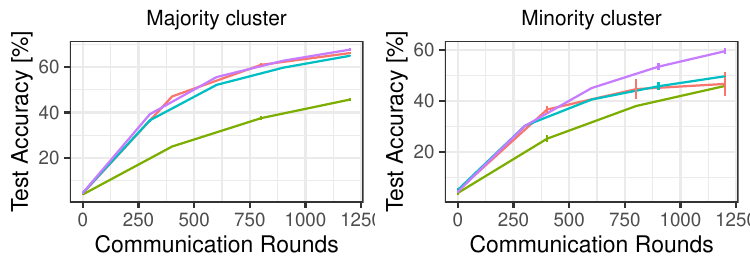}
		\caption{14:2 cluster configuration}
		\label{fig:acc_per_cluster_flickr_14_2}
	\end{subfigure}
	\caption{Average test accuracy for the nodes in the majority cluster (left) and those in the minority (right) obtained on \flickr ($\uparrow$ is better), for different cluster configurations.}
	\label{fig:acc_clust_flickr}
\end{figure}

\section{Experimental results for the \flickr dataset}
\label[appendix]{app:flickr_results}
In this section, we provide additional plots for the more challenging \flickr dataset.
\Cref{fig:acc_clust_flickr} shows the average test accuracy for the majority and minority clusters, for different cluster configurations and on the \flickr dataset.
\Cref{fig:acc_per_cluster_flickr_8_8} reveals that \sys for a 8:8 cluster configuration after $ T = 1200 $ reaches higher test accuracy than baselines, both for the minority and majority cluster.
\Cref{fig:acc_per_cluster_flickr_14_2} shows that for a 14:2 cluster configuration, \sys achieves comparable test accuracy for the majority cluster but shows a significant improvement in test accuracy for the minority cluster: 49.7\% for \EL against 59.6\% for \sys.

\begin{figure}[t]
	\includegraphics[width=\linewidth]{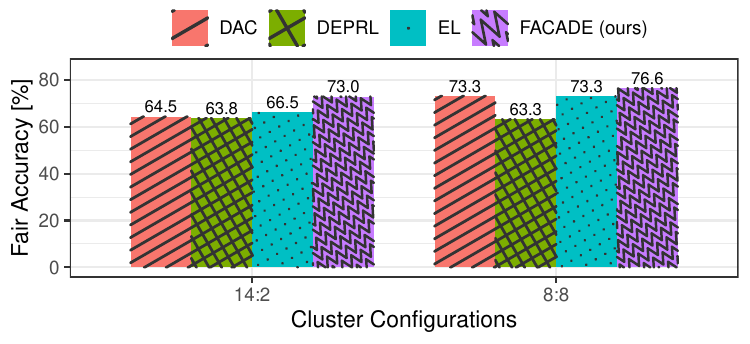}
	\caption{Highest observed fair accuracy for \flickr, for varying cluster configurations and algorithms ($\uparrow$ is better). To compute the fair accuracy, we use $\alpha = \nicefrac{2}{3}$.}
	\label{fig:fair_acc_flickr_barplot}
\end{figure}

\Cref{fig:fair_acc_flickr_barplot} shows the fair accuracy obtained on the \flickr dataset, for varying cluster configurations and algorithms.
In both cluster configurations, \sys achieves the highest fair accuracy.
This is in line with the results reported in~\Cref{sec:exp_fair_accuracy} for the other datasets.

\begin{figure}[t!]
	\begin{subfigure}{\columnwidth}
		\centering
		\includegraphics[width=.8\columnwidth]{figures/fairness_legend.pdf}
	\end{subfigure}
    \begin{subfigure}{0.99\linewidth} %
      \includegraphics[width=\linewidth]{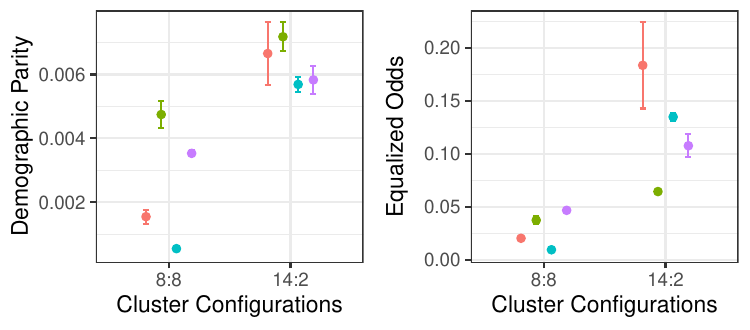}
    \end{subfigure}\hfill
\caption{Boxplot of \acl{DP} (left, $\downarrow$ is better) and \acl{EO} (right, $\downarrow$ is better) obtained on \flickr, for varying cluster configurations and algorithms.}
\label{fig:fairness_plots_flickr}
\end{figure}

\Cref{fig:fairness_plots_flickr} shows the \ac{DP} and \ac{EO} obtained on \flickr, for varying cluster configurations and algorithms.
\sys shows comparable \ac{DP} performance to \ac{EL} in an 8:8 cluster configuration.
However, \sys achieves a lower \ac{EO} score \ac{EO} than \EL and \dac in a 14:2 cluster configuration.
In the same configuration, \deprl achieves the lowest \ac{EO} score, which is in line with the results reported in~\Cref{sec:exp_fairness}.

\begin{table*}[t!]

\label{table:performance}

\begin{minipage}{\textwidth}
\centering
\caption{Summary of experimental results on the \cifar dataset.}
\label{tab:exp_summary_cifar10}

\begin{tabular}{l l c c c|c c|c}
\toprule
\textbf{\textsc{Config}} & \textbf{\textsc{Algorithm}} & \textbf{\textsc{Acc}\textsubscript{\textsc{maj}}$\uparrow$} & \textbf{\textsc{Acc}\textsubscript{\textsc{min}} $\uparrow$ } & \textbf{\textsc{Acc}\textsubscript{\textsc{all}} $\uparrow$ } & \textbf{\textsc{Demo. Par.} $\downarrow$} & \textbf{\textsc{Equ. Odds} $\downarrow$} & \textbf{\textsc{Acc}\textsubscript{\textsc{fair}} $\uparrow$ } \\
\midrule
\multirow{4}{*}{\rotatebox{90}{16:16}} & \EL & 64.03$\pm$\scriptsize{0.54} & 63.80$\pm$\scriptsize{0.42} & 63.91$\pm$\scriptsize{0.43} & 0.0032$\pm$\scriptsize{0.0007} & 0.0204$\pm$\scriptsize{0.0038} & 75.87 \\
                                       & \dac & 63.82$\pm$\scriptsize{1.34} & 63.79$\pm$\scriptsize{1.43} & 63.81$\pm$\scriptsize{0.31} & 0.0065$\pm$\scriptsize{0.0008} & 0.0402$\pm$\scriptsize{0.0057} & 75.86 \\
                                       & \deprl & 55.51$\pm$\scriptsize{0.34} & 55.50$\pm$\scriptsize{0.29} & 55.50$\pm$\scriptsize{0.26} & \textbf{0.0020$\pm$\scriptsize{0.0004}} & \textbf{0.0099$\pm$\scriptsize{0.0026}} & 70.33 \\
                                       & \textbf{FACADE} & \textbf{69.50$\pm$\scriptsize{0.32}} & \textbf{69.61$\pm$\scriptsize{0.20}} & \textbf{69.55$\pm$\scriptsize{0.25}} & 0.0060$\pm$\scriptsize{0.0015} & 0.0267$\pm$\scriptsize{0.0079} & \textbf{79.67} \\
\midrule
\multirow{4}{*}{\rotatebox{90}{24:8}}  & \EL & 69.13$\pm$\scriptsize{0.45} & 54.76$\pm$\scriptsize{0.59} & 65.53$\pm$\scriptsize{0.31} & 0.0103$\pm$\scriptsize{0.0012} & 0.1596$\pm$\scriptsize{0.0097} & 69.84 \\
                                       & \dac & 69.88$\pm$\scriptsize{0.28} & 51.30$\pm$\scriptsize{1.31} & 65.24$\pm$\scriptsize{0.19} & 0.0131$\pm$\scriptsize{0.0015} & 0.2067$\pm$\scriptsize{0.0175} & 67.53 \\
                                       & \deprl & 57.40$\pm$\scriptsize{0.13} & 54.21$\pm$\scriptsize{0.24} & 56.60$\pm$\scriptsize{0.15} & \textbf{0.0030$\pm$\scriptsize{0.0003}} & \textbf{0.0361$\pm$\scriptsize{0.0013}} & 69.48 \\
                                       & \textbf{FACADE} & \textbf{71.61$\pm$\scriptsize{0.27}} & \textbf{66.81$\pm$\scriptsize{0.34}} & \textbf{70.41$\pm$\scriptsize{0.29}} & 0.0079$\pm$\scriptsize{0.0025} & 0.0582$\pm$\scriptsize{0.0033} & \textbf{77.87} \\
\midrule
\multirow{4}{*}{\rotatebox{90}{30:2}}  & \EL & 71.99$\pm$\scriptsize{0.70} & 38.77$\pm$\scriptsize{0.62} & 69.91$\pm$\scriptsize{0.67} & 0.0306$\pm$\scriptsize{0.0014} & 0.3693$\pm$\scriptsize{0.0081} & 59.18 \\
                                       & \dac & 72.21$\pm$\scriptsize{0.24} & 34.94$\pm$\scriptsize{0.72} & 69.88$\pm$\scriptsize{0.25} & 0.0340$\pm$\scriptsize{0.0027} & 0.4143$\pm$\scriptsize{0.0075} & 56.63 \\
                                       & DePRL & 58.47$\pm$\scriptsize{0.59} & 52.56$\pm$\scriptsize{0.78} & 58.10$\pm$\scriptsize{0.57} & \textbf{0.0047$\pm$\scriptsize{0.0013}} & \textbf{0.0684$\pm$\scriptsize{0.0072}} & 68.37 \\
                                       & \textbf{FACADE} & \textbf{73.32$\pm$\scriptsize{0.15}} & \textbf{59.96$\pm$\scriptsize{0.72}} & \textbf{72.48$\pm$\scriptsize{0.19}} & 0.0086$\pm$\scriptsize{0.0026} & 0.1491$\pm$\scriptsize{0.0059} &\textbf{ 73.31} \\
\bottomrule
\end{tabular}

\end{minipage}\hfill
\vspace{1cm}
\begin{minipage}{\textwidth}
\centering
\caption{Summary of experimental results on the \imagenette dataset.}

\begin{tabular}{l l c c c|c c|c}
\toprule
 \textbf{\textsc{Config}} & \textbf{\textsc{Algorithm}} & \textbf{\textsc{Acc}\textsubscript{\textsc{maj}}$\uparrow$} & \textbf{\textsc{Acc}\textsubscript{\textsc{min}} $\uparrow$ } & \textbf{\textsc{Acc}\textsubscript{\textsc{all}} $\uparrow$ } & \textbf{\textsc{Demo. Par.} $\downarrow$} & \textbf{\textsc{Equ. Odds} $\downarrow$} & \textbf{\textsc{Acc}\textsubscript{\textsc{fair}} $\uparrow$ } \\
\midrule
\multirow{4}{*}{\rotatebox{90}{\textbf{12:12}}} & EL & 66.43$\pm$\scriptsize{0.56} & 66.85$\pm$\scriptsize{0.67} & 66.64$\pm$\scriptsize{0.59} & 0.0033$\pm$\scriptsize{0.0008} & 0.0208$\pm$\scriptsize{0.0035} & 77.62 \\
                                       & DAC & 65.73$\pm$\scriptsize{0.73} & 64.45$\pm$\scriptsize{0.54} & 65.09$\pm$\scriptsize{0.47} & \textbf{0.0054$\pm$\scriptsize{0.0004}} & \textbf{0.0286$\pm$\scriptsize{0.0036}} & 76.30 \\
                                       & DePRL & 43.14$\pm$\scriptsize{1.00} & 43.49$\pm$\scriptsize{1.20} & 43.31$\pm$\scriptsize{1.07} & 0.0078$\pm$\scriptsize{0.0016} & 0.0319$\pm$\scriptsize{0.0039} & 62.10 \\
                                       & \textbf{FACADE} & \textbf{68.18$\pm$\scriptsize{0.35}} & \textbf{68.59$\pm$\scriptsize{0.34}} & \textbf{68.39$\pm$\scriptsize{0.27}} & \textbf{0.0050$\pm$\scriptsize{0.0010}} & \textbf{0.0239$\pm$\scriptsize{0.0035}} & \textbf{78.78} \\
\midrule
\multirow{4}{*}{\rotatebox{90}{\textbf{16:8}}}  & EL & \textbf{69.69$\pm$\scriptsize{0.27}} & 60.21$\pm$\scriptsize{0.44} & 67.32$\pm$\scriptsize{0.29} & 0.0121$\pm$\scriptsize{0.0008} & 0.1061$\pm$\scriptsize{0.0050} & 73.48 \\
                                       & DAC & 68.55$\pm$\scriptsize{0.62} & 56.92$\pm$\scriptsize{1.32} & 65.64$\pm$\scriptsize{0.57} & 0.0136$\pm$\scriptsize{0.0019} & 0.1299$\pm$\scriptsize{0.0176} & 71.28 \\
                                       & DePRL & 43.40$\pm$\scriptsize{0.79} & 43.09$\pm$\scriptsize{1.22} & 43.33$\pm$\scriptsize{0.89} & 0.0096$\pm$\scriptsize{0.0025} & \textbf{0.0361$\pm$\scriptsize{0.0060}} & 62.06 \\
                                       & \textbf{FACADE} & \textbf{69.61$\pm$\scriptsize{0.37}} & \textbf{66.44$\pm$\scriptsize{0.19}} & \textbf{68.82$\pm$\scriptsize{0.30}} & \textbf{0.0054$\pm$\scriptsize{0.0011}} & \textbf{0.0397$\pm$\scriptsize{0.0014}} & \textbf{77.63} \\
\midrule
\multirow{4}{*}{\rotatebox{90}{\textbf{20:4}}}  & EL & \textbf{70.17$\pm$\scriptsize{0.28}} & 56.06$\pm$\scriptsize{0.56} & 67.81$\pm$\scriptsize{0.33} & 0.0186$\pm$\scriptsize{0.0006} & 0.1566$\pm$\scriptsize{0.0026} & 70.71 \\
                                       & DAC & 69.05$\pm$\scriptsize{0.74} & 50.93$\pm$\scriptsize{1.65} & 66.03$\pm$\scriptsize{0.45} & 0.0226$\pm$\scriptsize{0.0039} & 0.2009$\pm$\scriptsize{0.0249} & 67.29 \\
                                       & DePRL & 43.67$\pm$\scriptsize{1.01} & 42.64$\pm$\scriptsize{1.16} & 43.50$\pm$\scriptsize{1.02} & \textbf{0.0090$\pm$\scriptsize{0.0024}} & \textbf{0.0388$\pm$\scriptsize{0.0078}} & 61.76 \\
                                       & \textbf{FACADE} & \textbf{69.61$\pm$\scriptsize{0.46}} & \textbf{64.15$\pm$\scriptsize{0.39}} & \textbf{68.70$\pm$\scriptsize{0.42}} & \textbf{0.0064$\pm$\scriptsize{0.0015}} & 0.0630$\pm$\scriptsize{0.0057} & \textbf{76.10} \\
\bottomrule
\end{tabular}

\end{minipage}\hfill
\vspace{1cm}
\begin{minipage}{\textwidth}
\centering
\caption{Summary of experimental results on the \flickr dataset.}

\begin{tabular}{l l c c c|c c|c}
\toprule
 \textbf{\textsc{Config}} & \textbf{\textsc{Algorithm}} & \textbf{\textsc{Acc}\textsubscript{\textsc{maj}}$\uparrow$} & \textbf{\textsc{Acc}\textsubscript{\textsc{min}} $\uparrow$ } & \textbf{\textsc{Acc}\textsubscript{\textsc{all}} $\uparrow$ } & \textbf{\textsc{Demo. Par.} $\downarrow$} & \textbf{\textsc{Equ. Odds} $\downarrow$} & \textbf{\textsc{Acc}\textsubscript{\textsc{fair}} $\uparrow$ } \\
\midrule
\multirow{4}{*}{\rotatebox{90}{\textbf{8:8}}} & EL & 59.97$\pm$\scriptsize{0.23} & 59.92$\pm$\scriptsize{0.22} & 59.94$\pm$\scriptsize{0.19} & \textbf{0.0006$\pm$\scriptsize{0.0001}} & \textbf{0.0094$\pm$\scriptsize{0.0014}} & 73.28 \\
                                       & DAC & 60.56$\pm$\scriptsize{0.60} & 59.94$\pm$\scriptsize{0.32} & 60.25$\pm$\scriptsize{0.33} & 0.0016$\pm$\scriptsize{0.0002} & 0.0203$\pm$\scriptsize{0.0012} & 73.29 \\
                                       & DePRL & 44.92$\pm$\scriptsize{0.61} & 45.61$\pm$\scriptsize{1.23} & 45.26$\pm$\scriptsize{0.85} & 0.0047$\pm$\scriptsize{0.0004} & 0.0373$\pm$\scriptsize{0.0039} & 63.28 \\
                                       & \textbf{FACADE} & \textbf{65.50$\pm$\scriptsize{0.55}} & \textbf{64.92$\pm$\scriptsize{0.41}} & \textbf{65.21$\pm$\scriptsize{0.47}} & 0.0035$\pm$\scriptsize{0.0001} & 0.0467$\pm$\scriptsize{0.0018} & \textbf{76.62} \\
\midrule
\multirow{4}{*}{\rotatebox{90}{\textbf{14:2}}}  & EL & 64.92$\pm$\scriptsize{0.21} & 49.71$\pm$\scriptsize{0.20} & 63.02$\pm$\scriptsize{0.17} & \textbf{0.0057$\pm$\scriptsize{0.0002}} & 0.1349$\pm$\scriptsize{0.0034} & 66.47 \\
                                       & DAC & 66.11$\pm$\scriptsize{0.46} & 46.70$\pm$\scriptsize{4.75} & 63.68$\pm$\scriptsize{0.63} & 0.0067$\pm$\scriptsize{0.0010} & 0.1836$\pm$\scriptsize{0.0409} & 64.47 \\
                                       & DePRL & 45.69$\pm$\scriptsize{0.82} & 45.91$\pm$\scriptsize{0.79} & 45.72$\pm$\scriptsize{0.81} & 0.0072$\pm$\scriptsize{0.0005} & \textbf{0.0644$\pm$\scriptsize{0.0027}} & 63.80 \\
                                       & \textbf{FACADE} & \textbf{67.63$\pm$\scriptsize{0.49}} & \textbf{59.55$\pm$\scriptsize{1.06}} & \textbf{66.62$\pm$\scriptsize{0.56}} & \textbf{0.0058$\pm$\scriptsize{0.0004}} & 0.1077$\pm$\scriptsize{0.0109} & \textbf{73.03} \\
\bottomrule
\end{tabular}

\end{minipage}

\end{table*}

\section{Notes on \sys cluster assignment}
\label[appendix]{app:settlement_analysis}

\begin{figure}[t!]
    \begin{subfigure}{0.99\linewidth} %
      \includegraphics[width=\linewidth]{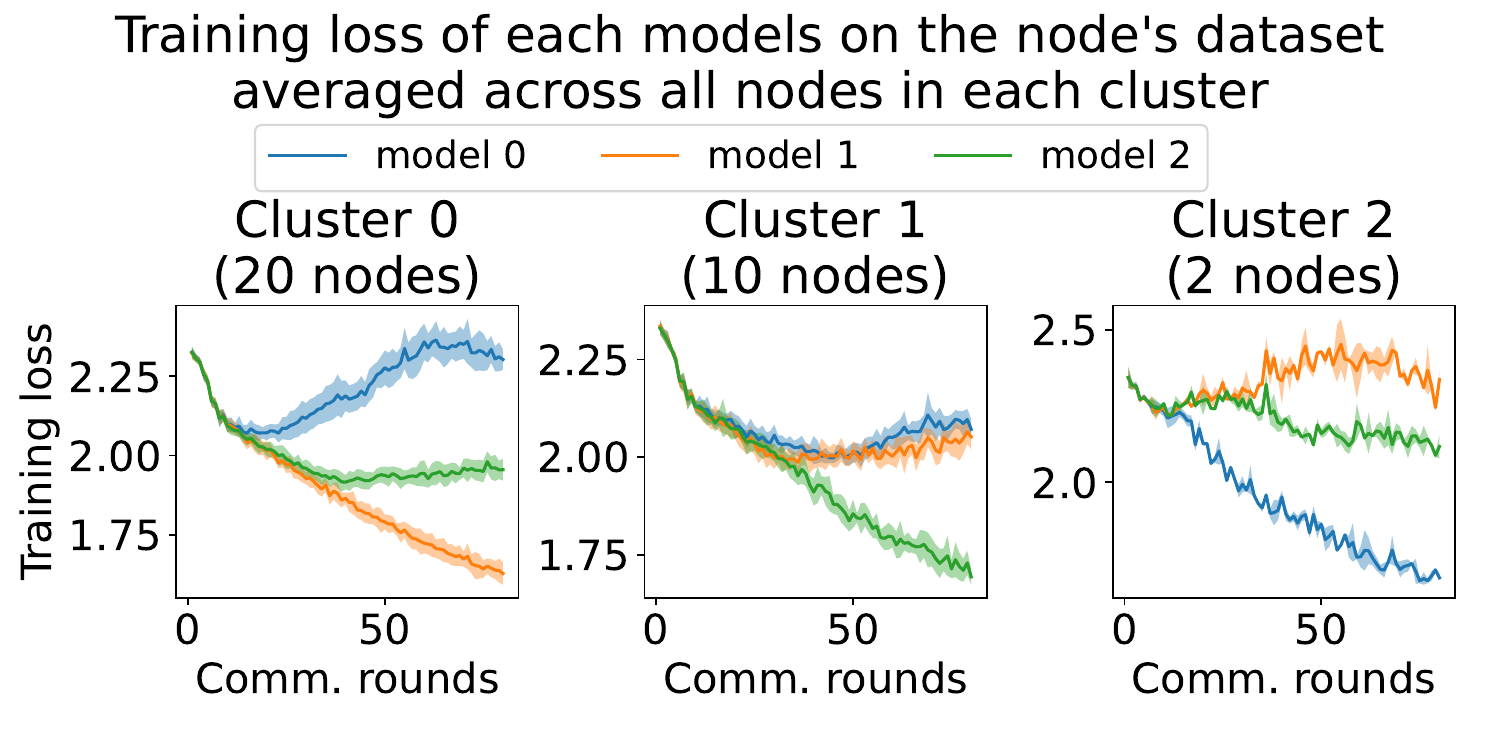}
      \caption{\sys did \emph{settle}}\label{subfig:all_mod_val}
    \end{subfigure}\hfill
    \vspace{0.5cm}
    \begin{subfigure}{0.99\linewidth} %
      \includegraphics[width=\linewidth]{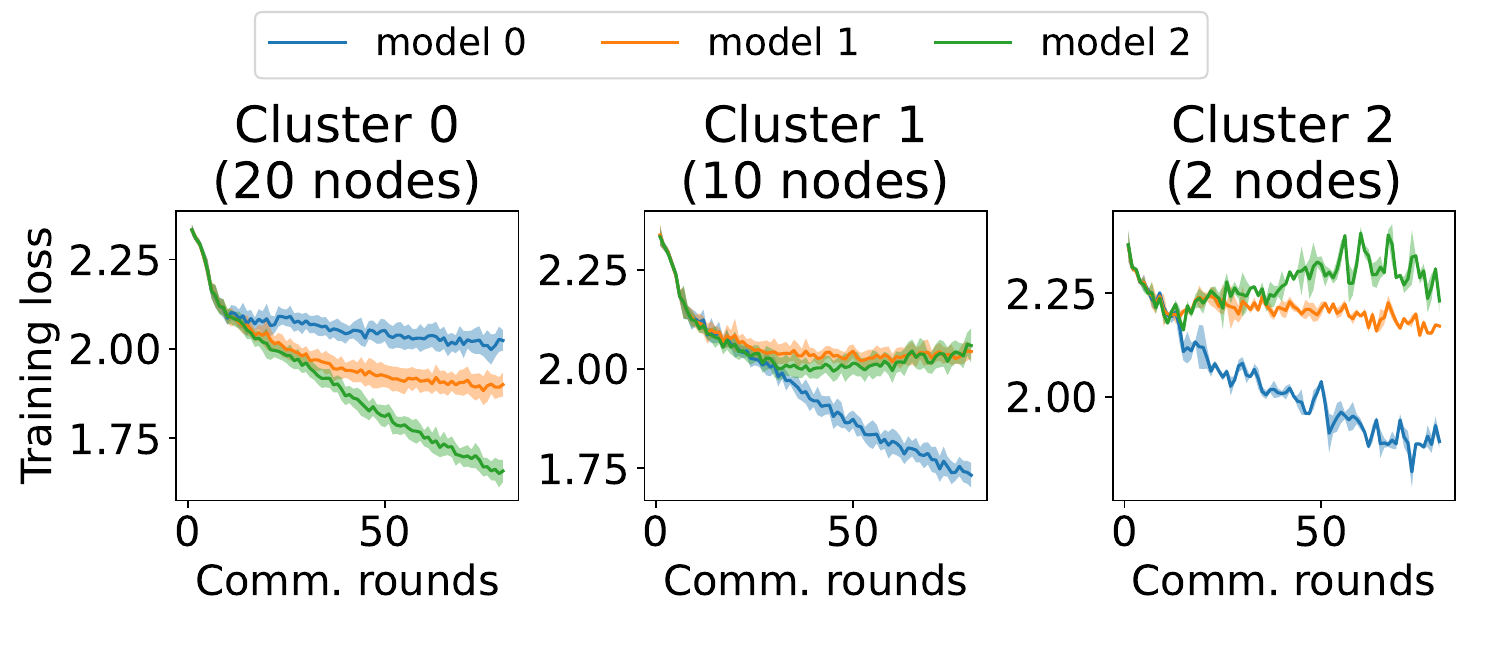}
      \caption{\sys did not \emph{settle}}\label{subfig:all_mo_val_fail}
    \end{subfigure}
\caption{The evolution of training loss ($\downarrow$ is better) on the shared core and each of the three model heads, averaged across nodes within the same cluster. The top plot illustrates a case where \sys successfully \emph{settled}, meaning all nodes within the same cluster favor the same model, and no nodes of another cluster picked it. The bottom plot shows a case where the algorithm did not \emph{settle}, resulting in all nodes from clusters 1 and 2 selecting and training the same model (model 0). We observe that model 1 is not selected at iteration 80, indicating it will no longer be chosen, as it will not be trained and thus will not improve on any distribution.}
\label{fig:settling}
\end{figure}

We mentioned in~\Cref{subsec:exp_settlement} that \sys sometimes fails to assign nodes to the correct cluster.
In this section, we provide additional insights into the concept of \emph{settlement} in our algorithm.
In \sys, a scenario can arise where one or more heads consistently outperform the others across all nodes.
When this happens, only the superior heads are selected, causing the other heads to be ignored and never be updated throughout the learning session.
This situation is referred to as the algorithm not \emph{settling}.
Figure~\ref{fig:settling} illustrates this behavior, with an example where \sys did settle (\Cref{subfig:all_mod_val} and another where it did not (\Cref{subfig:all_mo_val_fail}).

We empirically observed that the probability of not settling is higher when the cluster sizes are more imbalanced.
When the algorithm does not settle, it cannot fully exploit its potential and results in sub-optimal model utility and fairness.
However, it is important to note that not settling is not a catastrophic issue.
In such cases, performance in the worst case drops to the level of \EL, and a simple change of seeds is usually enough to achieve settlement.

To mitigate the risk of not settling, we employ several strategies.
First, with careful selection of model hyperparameters, we can reduce the likelihood of this occurrence.
Another effective strategy is to initiate the training with a few rounds of \EL, where all heads share the same weights before transitioning to independent parameters for each head.
This initial shared training phase is particularly crucial during the early stages of the algorithm when the models are still largely predicting randomly.
During this phase, one head can easily capture a better data representation and quickly outperform the others. 
By beginning with shared training, the core and heads develop a solid data representation foundation.
When the heads eventually train independently, it becomes easier for each to specialize in a specific cluster's data distribution, thus stabilizing the training process.
These techniques proved effective in stabilizing the training process and enhancing the overall performance of \sys.
By incorporating initial shared training, we significantly reduced the likelihood of the algorithm failing to settle, thereby maximizing its potential.

\section{Additional Experiments with Label Heterogeneity}
\label[appendix]{app:label_heterogeneity}

Our experiments in~\Cref{sec:eval} focus on evaluating \sys and baselines with feature heterogeneity by introducing a covariate shift through image rotations.
In line with related work~\cite{zec2022dac}, we also evaluate the performance of \sys and baselines with label heterogeneity where different nodes have different labels.
We use the \cifar dataset and consider a two-cluster setup where nodes in the first cluster get assigned images of vehicles (corresponding to the classes \textsc{car}, \textsc{plane}, etc.) and nodes in the second cluster images of animals (corresponding to the classes \textsc{dog}, \textsc{cat}, etc.).
Within each cluster, we uniformly divide these images across the nodes in the cluster.
Consistent with our experiments in~\Cref{sec:exp_test_accuracy}, we consider three cluster configurations with cluster size ratios of 16:16, 24:8, and 30:2.
Since there are more classes of animals as compared to vehicles in \cifar, this experiment also incorporates heterogeneity in terms of the number of samples at each node.

\begin{figure}[t]
	\centering
	\begin{subfigure}{\columnwidth}
		\centering
		\includegraphics[width=.8\columnwidth]{figures/acc_per_cluster_legend.pdf}
	\end{subfigure}
	\begin{subfigure}{\columnwidth}
		\centering
		\includegraphics[width=\columnwidth]{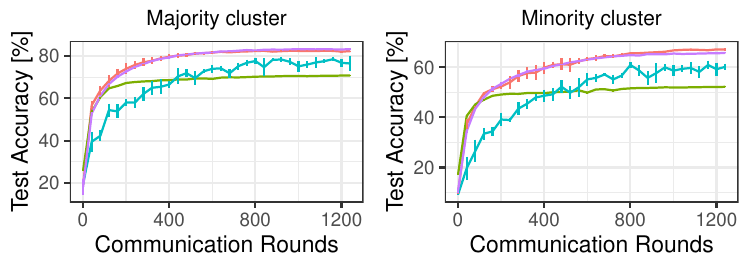}
		\caption{16:16 cluster configuration}
		\label{fig:acc_per_cluster_cifar10_16_16_labelshift}
	\end{subfigure}
	\begin{subfigure}{\columnwidth}
		\centering
		\includegraphics[width=\columnwidth]{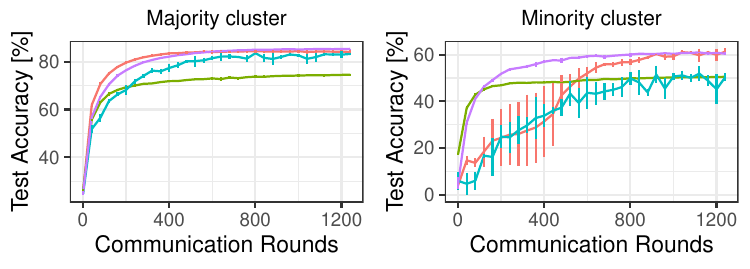}
		\caption{24:8 cluster configuration}
		\label{fig:acc_per_cluster_cifar10_24_8_labelshift}
	\end{subfigure}
	\begin{subfigure}{\columnwidth}
		\centering
		\includegraphics[width=\columnwidth]{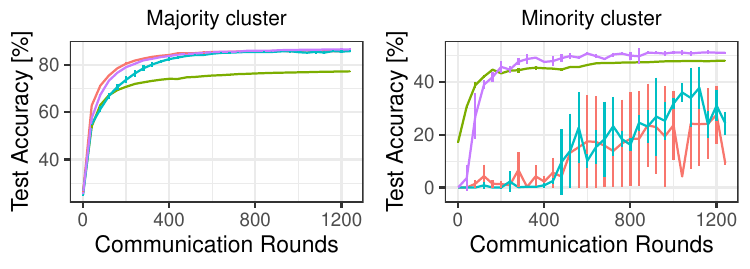}
		\caption{30:2 cluster configuration}
		\label{fig:acc_per_cluster_cifar10_30_2_labelshift}
	\end{subfigure}%
	\caption{Average test accuracy for the nodes in the majority cluster (left) and those in the minority (right) obtained on \cifar ($\uparrow$ is better), for different cluster configurations and with label heterogeneity.}
	\label{fig:acc_clust_cifar_labelshift}
\end{figure}

\Cref{fig:acc_clust_cifar_labelshift} shows the average test accuracy for each cluster for the \cifar dataset as model training progresses for the three considered cluster configurations.
The test accuracy of the majority and minority clusters is shown in the left and right columns, respectively.
In all cluster configurations, we observe equal performance of \sys and \dac.
Additionally, \EL reaches similar performance to both \sys and \dac in the 30:2 cluster configuration (\Cref{fig:acc_per_cluster_cifar10_30_2_labelshift}, left) but converges slower.
While \dac shows competitive performance for nodes in the minority cluster in the 16:16 setting, its achieved accuracy drops significantly when cluster sizes are very imbalanced (\Cref{fig:acc_per_cluster_cifar10_30_2_labelshift}, right).
Both \EL and \dac exhibit unpredictable performance in this setting and fail to converge.
In this scenario, \sys achieves the highest accuracy for nodes in the minority cluster: 51.2\% test accuracy after \num{1200} communication rounds.
In contrast, \deprl achieves 48.1\% test accuracy after \num{1200} communication rounds.
Thus, \sys attains high test accuracies for nodes in all clusters and across all cluster configurations.

\begin{figure}[t]
	\includegraphics[width=\linewidth]{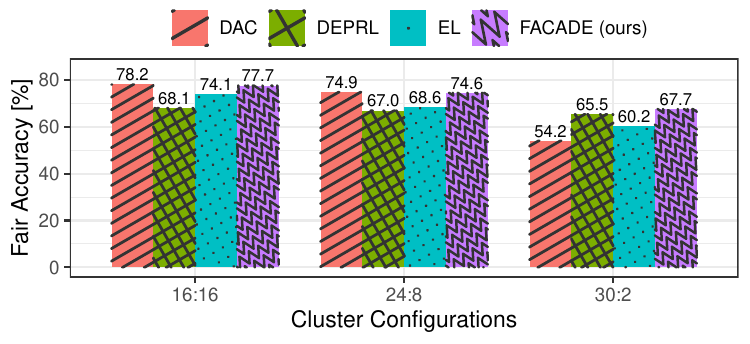}
	\caption{Highest observed fair accuracy for \cifar, for varying cluster configurations and algorithms ($\uparrow$ is better) and with label heterogeneity. To compute the fair accuracy, we use $\alpha = \nicefrac{2}{3}$.}
	\label{fig:fair_acc_cifar10_barplot_labelshift}
\end{figure}

\Cref{fig:fair_acc_cifar10_barplot_labelshift} shows the highest obtained fair accuracy for \cifar, for the different cluster configurations, baselines, and with label heterogeneity.
In line with our other experiments, we use $\alpha = \nicefrac{2}{3}$ to compute these fair accuracies.
\sys reaches competitive fair accuracy for the 16:16 and 24:8 cluster configurations.
\sys, however, achieves the highest fair accuracies for the 30:2 cluster configuration: 67.7\% compared to 65.5\% for the \deprl baseline.

\begin{figure}[t]
	\includegraphics[width=\linewidth]{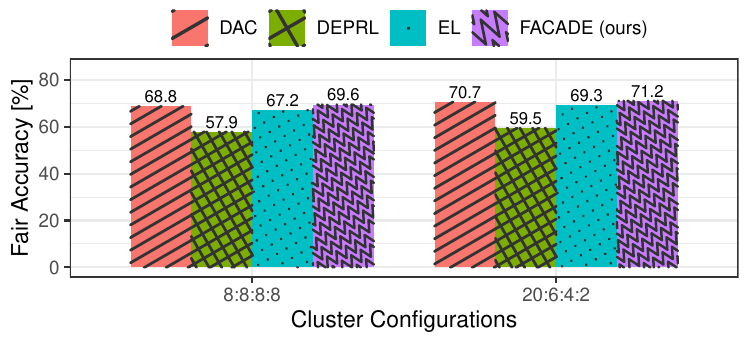}
	\caption{Highest observed fair accuracy for \cifar, for varying cluster configurations and algorithms ($\uparrow$ is better) and with feature heterogeneity by appyling color filters. To compute the fair accuracy, we use $\alpha = \nicefrac{2}{3}$.}
	\label{fig:fair_acc_cifar10_barplot_colorshift}
\end{figure}

\section{Additional Experiments with Color Shifting}
\label[appendix]{app:exp_color_shift}

We next experiment with \cifar and a different source of feature heterogeneity than rotation by applying three color filters to training images and leaving the images in one cluster untouched, effectively creating four clusters.
We apply the grayscale, sepia, and high saturation filters to each cluster and consider two cluster configurations: 8:8:8:8 (balanced) and 20:6:4:2 (imbalanced).

\begin{figure*}[t!]
    \centering
    \begin{subfigure}{\columnwidth}
		\centering
		\includegraphics[width=.8\columnwidth]{figures/acc_per_cluster_legend.pdf}
	\end{subfigure}
    \vspace{0.1cm}
    \begin{subfigure}{0.99\linewidth} %
      \includegraphics[width=\linewidth]{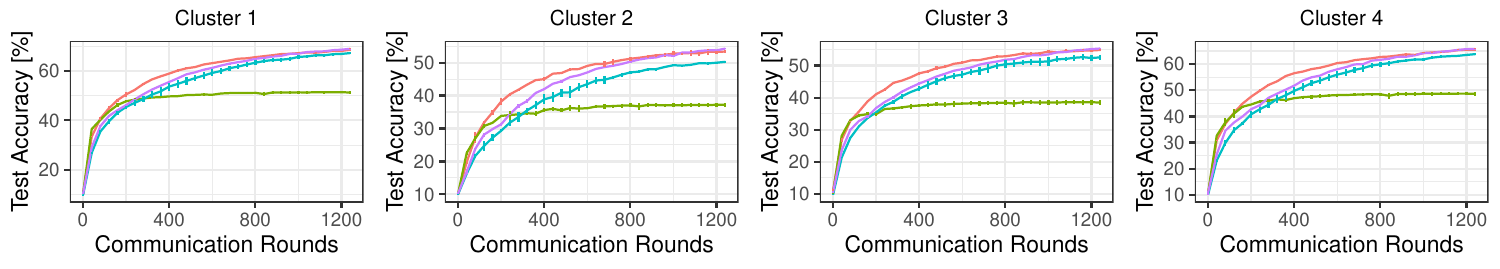}
      \caption{8:8:8:8 cluster configuration}\label{fig:acc_per_cluster_cifar10_8_8_8_8_colorshift}
    \end{subfigure}\hfill
    \vspace{0.5cm}
    \begin{subfigure}{0.99\linewidth} %
      \includegraphics[width=\linewidth]{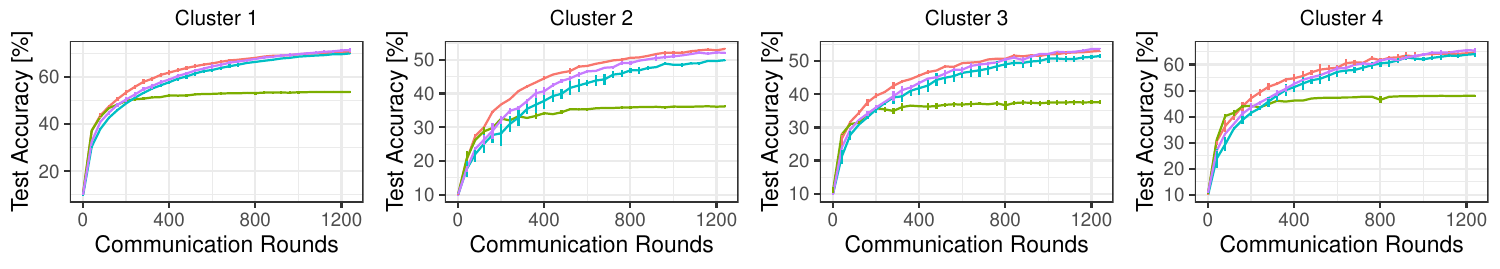}
      \caption{20:6:4:2 cluster configuration}\label{subfig:acc_per_cluster_cifar10_20_6_4_2_colorshift}
    \end{subfigure}
\caption{Average test accuracy for the nodes in different clusters obtained on \cifar ($\uparrow$ is better), for different cluster configurations and with feature heterogeneity through color shifting.}
\label{fig:acc_clust_cifar_colorshift}
\end{figure*}

\Cref{fig:acc_clust_cifar_colorshift} shows the average test accuracy for each cluster for the \cifar dataset as model training progresses for the two considered cluster configurations.
After \num{1200} communication rounds, \sys reaches the highest test accuracy for all cluster configurations and settings, except in the presence of the sepia filter in the 20:6:4:2 configuration.
However, the difference in test accuracy compared to \dac is small: around a half percent point.
In line with our observations in~\Cref{sec:eval}, the accuracy increase of \deprl stalls after a few hundred rounds.
We also observe that the convergence of \sys is slightly slower than \dac, yet both approaches attain comparable final test accuracy.

\Cref{fig:acc_clust_cifar_colorshift} shows that \sys and baselines, compared to~\Cref{fig:acc_clust_cifar}, have relatively good performance across all clusters and cluster configurations.
We believe this is because cluster-based training when applying the chosen color filters is an easier problem than when images are rotated.
These color filters result in feature variations that are less disruptive to the underlying spatial structure of the data compared to rotations.
Whereas rotations fundamentally alter the orientation of key features, requiring the model to learn orientation-invariant representations, color filters merely modify the intensity or color channels.
The model can adapt to this more easily using standard feature extraction layers.
This is also evident from the fact that the traditional baseline \ac{EL} reaches a decent accuracy across all clusters and configurations.

\Cref{fig:fair_acc_cifar10_barplot_colorshift} shows the highest obtained fair accuracy for \cifar, for the different cluster configurations, baselines, and with feature heterogeneity by applying color filters.
With both cluster configurations, \sys achieves the highest fair accuracy.
In the balanced setting (8:8:8:8 cluster configuration), \sys obtains 69.6\% fair accuracy compared to 68.8\% for \dac, the best-performing baseline.
In the imbalanced setting (20:6:4:2 cluster configuration), \sys obtains 71.2\% fair accuracy compared to 70.7\% for \dac.
In summary, \sys outperforms the baselines in achieving a good model while minimizing the accuracy difference across clusters.

\section{\sys and Privacy}
\label[appendix]{app:privacy}

While our primary focus in this work is on improving fairness, we acknowledge that privacy is a critical consideration for \ac{DL} systems~\cite{pasquini2023security,biswas2024beyond}.
\ac{DL} offers an additional level of privacy protection compared to centralized approaches where data is shared directly.
However, parameter sharing in \ac{DL} could still leak information about training data, particularly in scenarios with highly imbalanced clusters or when minority clusters are small.
Addressing privacy concerns is an important direction for future work.
Techniques such as differential privacy~\cite{lin2022towards,biswas2024lowcost} or secure aggregation~\cite{mansouri2023sok} could be incorporated into \sys to mitigate such risks.
We leave studying the privacy risks of fairness-enhancing techniques such as \sys for future work.

\end{document}